\PassOptionsToPackage{table}{xcolor}
\documentclass[dvipsnames]{article} %
\usepackage{colm2024_conference}

\usepackage{booktabs}
\usepackage{float}
\usepackage{enumitem}
\usepackage{wrapfig}
\usepackage{graphicx}
\graphicspath{{figures/}}
\usepackage[misc]{ifsym}
\usepackage{multicol}
\usepackage{amssymb}
\usepackage{microtype}
\usepackage{amsmath}
\usepackage{colortbl}
\usepackage[utf8]{inputenc}
\usepackage[T1]{fontenc}
\usepackage{iftex}
\ifXeTeX
  \usepackage{newunicodechar}
  \newunicodechar{’}{'}
  \newunicodechar{—}{---}
\fi
\definecolor{lightgray}{rgb}{0.9,0.9,0.9}
\usepackage{caption}
\usepackage{subcaption}
\usepackage{setspace}
\usepackage{url}
\usepackage{multirow}
\usepackage{tabularx}
\newcolumntype{Y}{>{\centering\arraybackslash}X}
\usepackage{makecell}
\usepackage{siunitx}
\usepackage{listings}
\usepackage[raster,skins, most]{tcolorbox}
\usepackage{xltabular}
\usepackage{adjustbox}
\usepackage{xurl}
\usepackage{rotating}
\usepackage[normalem]{ulem}
\usepackage[subfigure]{tocloft}
\usepackage{needspace}
\ifPDFTeX\usepackage{fontawesome}\fi
\usepackage{array}
\usepackage[table]{xcolor}
\usepackage{placeins}  %
\usepackage{needspace} 

\useunder{\uline}{\ul}{}

\usepackage{amsmath,amsfonts,bm}
\usepackage{pifont}

\def\eqref#1{equation~\ref{#1}}

\def\1{\bm{1}}

\DeclareMathAlphabet{\mathsfit}{\encodingdefault}{\sfdefault}{m}{sl}
\SetMathAlphabet{\mathsfit}{bold}{\encodingdefault}{\sfdefault}{bx}{n}

\definecolor{uclablue}{RGB}{159, 195, 224}

\definecolor{aliceblue}{RGB}{255, 238, 241}

\definecolor{babygreen}{rgb}{0.85, 0.97, 0.85}

\definecolor{deepblue}{RGB}{48, 58, 82}
\definecolor{deepPurple}{HTML}{330066}
\definecolor{gcuarow}{RGB}{230, 245, 255}
\definecolor{baselineScore}{HTML}{E4E4FB}
\definecolor{ogcuaScore}{HTML}{CFCDFD}
\definecolor{realdeviceHighlight}{HTML}{CFCDFD}

\definecolor{uclablue_old}{rgb}{0.15, 0.45, 0.68}
\hypersetup{
    breaklinks,
    citecolor=uclablue_old,
    colorlinks=true,
}
\newtcbtheorem{promptbox}{Prompt}{
    breakable,
    colframe=deepPurple!70,
}{box}

\newtcolorbox{mybox}[2][]
  {colback = black!5!white, colframe = black!75!black, fonttitle = \bfseries,
    colbacktitle = black!100!black, enhanced, before upper={\fontsize{8}{11}\obeyspaces\obeylines\selectfont}, fontupper=\selectfont,
    attach boxed title to top left={yshift=-2.2mm,xshift=4mm},
    title=#2,#1}

\newcommand{\projname}{Open Generalist CUA}
\newcommand{\projshort}{OpenGCUA}
\newcommand{\qwenui}{Qwen-UI-Agent}
\newcommand{\sdkname}{Cross-Device SDK}
\newcommand{\qwenmax}{Qwen3.8-Max}
\newcommand{\qwenflash}{Qwen3.8-Flash}
\newcommand{\osworldtwo}{OSWorld-v2}

\newcommand{\websiteicon}{%
  \raisebox{-0.12em}{%
    \tikz[x=1em,y=1em,line width=0.4pt,draw=black]{
      \draw (0,0) circle (0.45);
      \draw (0,0) ellipse (0.20 and 0.45);
      \draw (-0.45,0) -- (0.45,0);
    }%
  }%
}

\title{
    OpenGCUA: Toward a Fully Open Stack for Generalist Computer Use
}

\author{%
  {\fontsize{9.4pt}{12pt}\selectfont
  \begin{tabular}{@{}c@{}}
  Hanzhang Zhou\textsuperscript{*},\hspace{0.25em}%
  Long Li\textsuperscript{*},\hspace{0.25em}%
  Quyu Kong\textsuperscript{*},\hspace{0.25em}%
  Lei Wang\textsuperscript{*},\hspace{0.25em}%
  Panrong Tong\textsuperscript{*},\hspace{0.25em}%
  Long Chen,\hspace{0.25em}%
  Tianyu Xia,\hspace{0.25em}%
  Gongjie Zhang,\\[1.2pt]
  Xu Zhang,\hspace{0.25em}%
  Chen Liu,\hspace{0.25em}%
  Jianan Zhang,\hspace{0.25em}%
  Chenglin Cai,\hspace{0.25em}%
  Zhengwen Zeng,\hspace{0.25em}%
  Steven Hoi,\hspace{0.25em}%
  Yue Wang\textsuperscript{\ensuremath{\dagger}}\\[0.65em]
  \textbf{Alibaba Token Hub, Alibaba Group}
  \end{tabular}}%
}

\begin{document}

\vspace*{-15pt}
\maketitle
\enlargethispage{5pt}

\begingroup
\makeatletter
\renewcommand{\@makefntext}[1]{\noindent #1}
\makeatother
\renewcommand{\thefootnote}{}
\footnotetext{%
\textsuperscript{*}\,Equal contribution.\qquad
\textsuperscript{\ensuremath{\dagger}}\,Corresponding author.\\[1.5pt]
{The MAI-UI Team presents the first release of \projshort{}; further project releases will follow.}}
\endgroup

\vspace{-0.35em}
\begin{abstract}
\vspace{-0.3em}
Computer use has reached a turning point: the focus is expanding from GUI-only interaction to systems that complete tasks on users' computers through visual and programmatic interfaces.
We define a \emph{generalist computer-use agent} (GCUA) as a system in which the model chooses how to observe, act, and organize its work across visual and programmatic interfaces.
Yet frontier implementations, such as GPT-6 Astra's computer-use system, remain largely closed, limiting systematic study of their design.
We introduce \textbf{\projname{}} (OpenGCUA), an ongoing project to develop and open the full stack for GCUAs with the community.\\[2pt]
This report presents the first release of \projname{}.
\emph{Code execution} is a key feature of our agent design: it lets intermediate results determine subsequent operations and enables models to apply their existing programming capabilities directly to computer interaction.
We also propose \emph{harness-to-model decision transfer}, moving selected observations and task-state decisions from fixed harness rules to the model. 
On \osworldtwo{}, our full configuration improves task scores across all eight evaluated models, including GPT-6 Astra and Claude Opus 5, raising mean partial score by \textbf{+19.6} points and reducing reported steps per task by an average of \textbf{53.0\%} relative to batched GUI-only execution.
We release our implementation and prompts, evaluation trajectories from all eight models, and preliminary versions of a \emph{real-device harness} and a \emph{cross-device SDK}.
The harness addresses real-device challenges not fully captured by benchmarks, while the SDK provides a common programming interface across computers, mobiles and browsers.
Across four real macOS tasks under our real-device harness, selected \qwenmax{} runs use \textbf{19.4} times as many reported reasoning tokens as GPT-6 Astra, highlighting substantial differences between models in real-device execution efficiency.

\end{abstract}

\begin{figure}[H]
\vspace{-0.5em}
    \centering
    \includegraphics[width=0.99\linewidth]{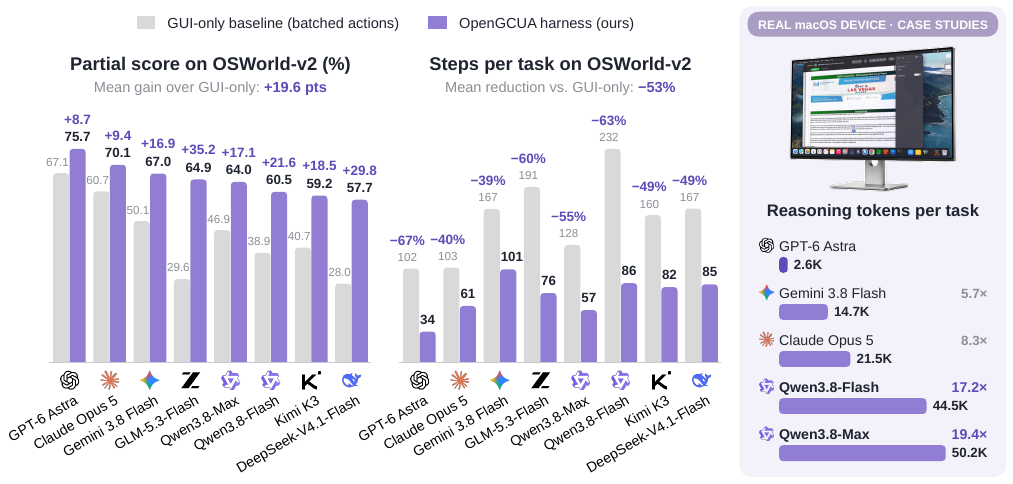}
    \vspace{-0.5em}
    \caption{On the offline set of \osworldtwo{}, \projshort{} improves all eight frontier models, raising mean partial score by \textbf{19.6} points and reducing steps per task by \textbf{53\%}. On real-device, the gap between models remains large: \qwenmax{} uses \textbf{19.4$\times$} GPT-6 Astra's reasoning tokens but completes fewer tasks.}
    \label{fig:teaser}
\end{figure}

\renewcommand{\thefootnote}{\arabic{footnote}}

\clearpage
\tableofcontents
\clearpage
\section{Introduction}
\vspace{-0.5em}
Computer use has moved to the center of attention. The GPT-6 Astra release places strong emphasis on computer use, with extensive demonstrations offering an exciting vision of what computer-use agents could make possible \citep{openai2026gpt6astra}. 
Much of the literature studies computer-use agents (CUAs) as GUI-only agents, asking how well models can complete computer tasks through screenshot-based observation and mouse and keyboard actions \citep{uitars, lu2026qwen}.
Recent frontier systems, however, combine GUI interaction with shell commands, accessibility-tree access, and other tools, and OpenAI now recommends code execution, writing GUI operations as code and combining them with programmatic operations in a single model output \citep{openai2026cuaguide}. 
These systems also target tasks on users' own computers, where shared control, interaction with background windows, and evolving requests introduce challenges not fully captured by benchmarks.
The question has changed, from how well a model operates a GUI interface to how well a system completes work on a real computer by whatever means works. We take this as \emph{a turning point} in computer-use research, motivating a clearer definition of these broader agents and a systematic study of their design.

We therefore adopt the term \textbf{generalist computer-use agent} (GCUA) to characterize this broader class of systems, following the system-level perspective of Agents' Last Exam \citep{agentslastexam}. GPT-6 Astra's computer-use system is a representative GCUA. We define a GCUA as \emph{an agent system designed to complete diverse computer tasks through both visual interaction and programmatic access, in which the model decides how to observe, act, and organize its work as the task requires}. Programmatic access includes shell commands, accessibility APIs, and application or service APIs where available. 
Essentially, a GCUA chooses how to use the computer to complete a task, rather than being restricted to GUI interaction. Our previous work, \qwenui{}, began exploring this direction through a unified GUI and CLI action space with batched actions \citep{qwenuiagent}. Within our formulation, such GUI+CLI hybrid CUA is a special case of GCUA (Section~\ref{sec:gcua}).

How to design such systems is an open question.
A GCUA raises design questions that go beyond those of GUI-only agents: which interfaces to give the model, how to support code execution, which decisions to leave to the model and which to the harness, and what changes when the agent runs on a user's own computer. 
However, frontier systems such as GPT-6 Astra remain largely closed \citep{openai2026codexcu}.
Their model weights, full prompting setups, real-device harnesses, device-control SDKs, and execution trajectories are not available.
This limits the community's ability to reproduce system behavior, compare alternative designs, and investigate how design choices and execution mechanisms affect task completion and cost, particularly under real-device conditions.

\begin{figure}[!t]
    \centering
    \includegraphics[width=0.95\linewidth]{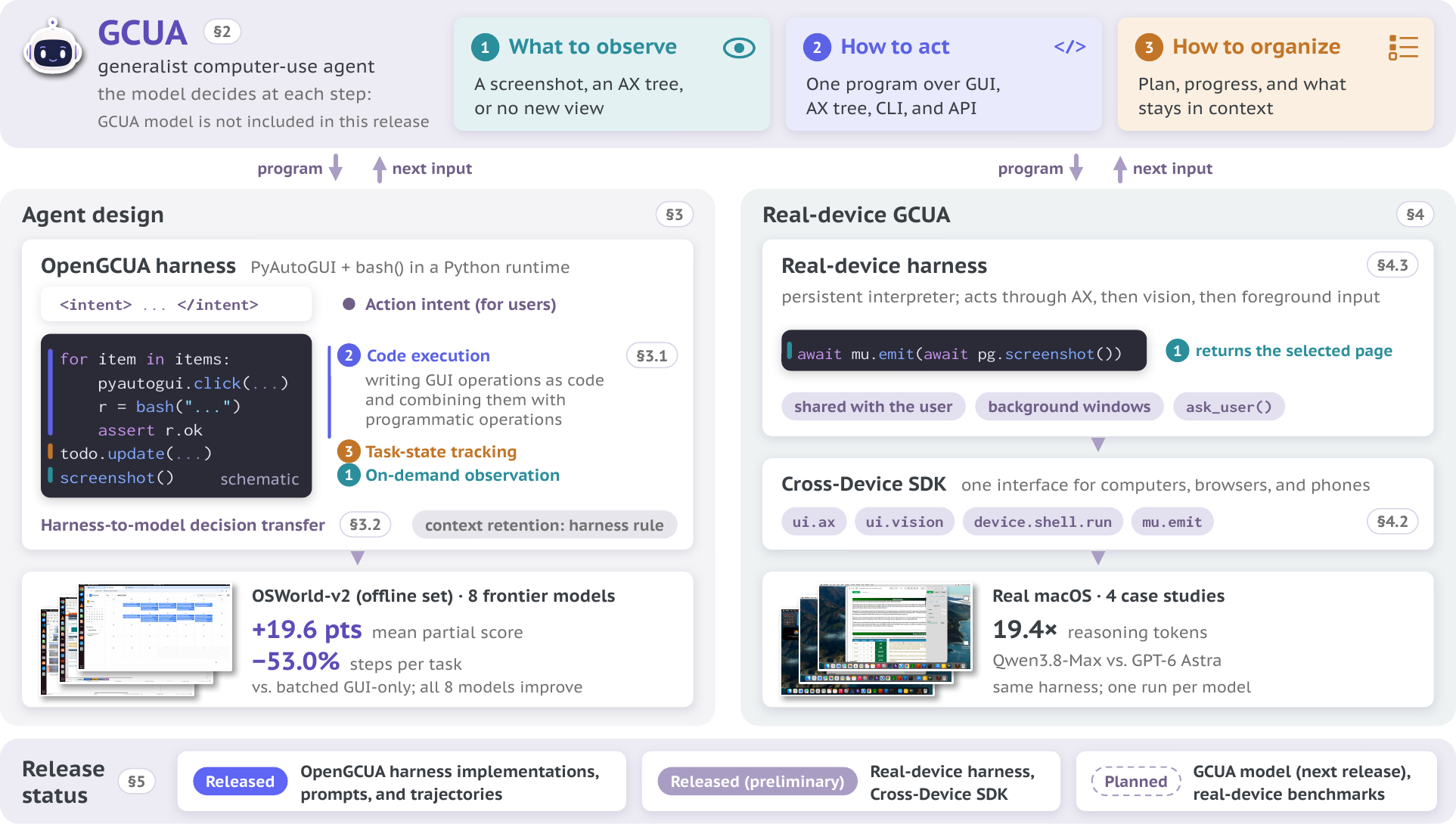}
    \caption{\textbf{Overview of \projname{}.}}
    \label{fig:ogcua-overview}
    \vspace{-1em}
\end{figure}

We therefore introduce \projname{} (\projshort{}), an ongoing project to develop and open the full stack for generalist computer use with the community (Figure~\ref{fig:ogcua-overview}):
\textbf{(1) Model.} Develop a model that unifies computer and browser use, with code execution and harness-to-model decision transfer central to its agent design.
\textbf{(2) System.} A real-device harness that runs on users' real machines, and lessons we learned in moving from benchmark harnesses to real-device deployment. A cross-device SDK that gives agent programs one interface to computers, browsers, and phones, supporting tasks that span multiple devices.
\textbf{(3) Trajectories.} Release trajectories from evaluations of frontier models such as GPT-6 Astra to support analysis of code-execution and harness-to-model decision transfer patterns.
\textbf{(4) Efficiency.} Systematically study and improve GCUA execution efficiency through code-execution, concise model reasoning, and harness optimization, while exploring broader system-level improvements.
\textbf{(5) Evaluation.} Develop evaluation environments and benchmarks on real devices, particularly on macOS, to assess agents' ability to complete tasks under real-world operating conditions.

This report presents the first release of the \projshort{} project. The release does not yet include a trained model. The report covers agent design and evaluation, execution trajectories from frontier models, and preliminary versions of the real-device harness and cross-device SDK, together with a roadmap for subsequent releases. Our main contributions are as follows.
\begin{itemize}[leftmargin=*]
    \item \textbf{GCUA generalizes existing GUI, CLI, and hybrid agents.}
    We define a GCUA by three decisions that its model makes at each step: what to observe, how to act, and how to organize its work. We show that GUI agents, CLI agents, multi-agent hybrids, and GUI+CLI agents such as \qwenui{} each fix some of these decisions in advance, and are special cases of GCUA (Section~\ref{sec:gcua}).

    \item \textbf{GCUA outperforms GUI-only CUA on every frontier model we test.}
    We implement our GCUA agent design into the \projshort{} harness, which lets existing models use code execution and harness-to-model decision transfer. We test it with eight frontier models, including GPT-6 Astra and Claude Opus 5. On the offline set of \osworldtwo{}, the full GCUA setting raises the mean partial score by \textbf{19.6 points}, cuts steps per task by \textbf{53.0\%}, and narrows the score gap between the strongest and weakest models by more than half. We release the implementation, prompts, and \textbf{evaluation trajectories of all eight frontier models} (Section~\ref{sec:model}).

    \item \textbf{Code execution adds value beyond shell access.}
    With the same GUI and shell access,  writing GUI operations as code and combining them with programmatic operations adds \textbf{6.1 points} to the mean partial score and cuts steps per task by \textbf{14.8\%} against the GUI+CLI baseline. 
    Code execution offers two advantages: intermediate results can decide the next operation without another model call, and models can bring their existing programming ability directly to computer interaction. We further identify five recurring patterns of code execution in our case study (Section~\ref{sec:code-exec}).

    \item \textbf{Harness control should move into the model as models improve.}
    Harnesses fix many workflow decisions by rule, such as taking a screenshot after every step. We propose \emph{harness-to-model decision transfer}, which delegates these harness decisions to the model. In the current implementation, the model decides when to request screenshots and how to track task state. It also states each program's purpose before execution. On-demand observation alone increases cache hit rates by 10.5 and 14.4 percentage points for two models without lowering task scores (Section~\ref{sec:selfdirected}).

    \item \textbf{On real devices, a large gap lies in the model.}
    We release a preliminary \emph{real-device harness} for users' own computers, addressing challenges unseen in benchmarks, such as sharing the machine with its user and acting on background windows. With the harness held fixed, \qwenmax{} uses \textbf{19.4 times} as many reasoning tokens as GPT-6 Astra on four real macOS tasks, and \textbf{5.7 times} as many output tokens in total. Case studies trace this gap to model behavior: large reasoning cost for simple operations, difficulty adapting to real-device conditions and a gap in code execution capability (Section~\ref{sec:realmachine}).

    \item \textbf{One Programming Interface for Computers, Browsers, and Phones.}
    The preliminary \emph{\sdkname{}} binds every operation to an explicit device, application, window, or tab, keeps accessibility and vision as separate ways to observe and act, and allows one program to coordinate several devices within a task (Section~\ref{sec:sdk}).

\end{itemize}

This release does not mark the project's completion, and the harness and SDK remain preliminary. 
Section~\ref{sec:roadmap} outlines our research plans and release milestones for \projname{}.

\section{Generalist CUA: Definition and Positioning}
\label{sec:gcua}
\vspace{-0.5em}

\subsection{Definition}
\label{sec:gcua-def}
Agents' Last Exam takes a system-level view and decomposes an agent into five layers: the model (Brain), GUI perception (Eyes), orchestration (Body), tools (Hands), and a runtime (Feet)~\citep{agentslastexam}. Under this view, a CLI agent lacks GUI perception, a GUI agent lacks most of the remaining layers, and a generalist agent spans all five.
Having all five layers, however, only means an agent can act through both the GUI and the CLI; it does not mean that the agent chooses well between them. 
At a given step, an agent may click through a dozen screens to do what a single shell command would do, or run a script when it first needs to look at the screen. The layer view cannot tell such an agent apart from one that picks the right way to act at each step.
In this report, \textbf{we therefore define a generalist CUA (GCUA) by the decisions it makes at each step, not by the components it contains}.

A GCUA makes three decisions at each step: what to observe, how to act, and how to organize.
\begin{itemize}[leftmargin=*]
\vspace{-0.5em}
\item \textbf{What to Observe.} The observation can be a screenshot, an accessibility tree (AX Tree), the output of a command, the response of an API, or any combination of these. The model decides which it needs, including whether it needs a new screenshot at all.
\item \textbf{How to Act.} The action can be a GUI operation at a screen coordinate, an operation on an accessibility element, a shell command, an API call, or a program that combines several of these. The model decides which fits the step.
\item \textbf{How to Organize.} Across steps, the agent keeps a plan, tracks what has been done, and decides which past observations stay in context. Most harnesses fix these as rules set by the developer. In a GCUA, the model makes these decisions as the task requires.
\end{itemize}

\paragraph{Four Interfaces.}
The first two decisions range over four interfaces to the computer, and each offers both a way to observe and a way to act. 
\begin{itemize}[leftmargin=*]
\vspace{-0.5em}
\item \textbf{GUI.} The agent sees a screenshot and clicks, drags, scrolls, or types at a screen coordinate. This is the interface that always works, since every application has pixels.
\item \textbf{Accessibility Tree.} The agent reads the elements of a window, with their roles, names, and values, and operates on an element directly. It needs no coordinate, does not bring the window to the front, and never moves the user's pointer on real devices.
\item \textbf{CLI.} The agent runs shell commands and reads their output. This is the natural route for files, data, and software.
\item \textbf{API.} The agent calls external services, such as web search, and reads their responses. This reaches what neither the screen nor the shell can.
\end{itemize}

Each interface offers both a way to observe and a way to act, and each has its own strength. No single interface is best for every step, so what matters is choosing the right one at each step and combining them when a step needs more than one. 

Putting these together yields the definition used throughout the report: \emph{A GCUA is an agent system designed to complete diverse computer tasks through both visual interaction and programmatic access, in which the model decides how to observe, act, and organize its work as the task requires}.
This definition refines that of ALE rather than replacing it. Visual interaction and programmatic access require all five layers, and our definition further asks who makes the decisions within them.

\paragraph{Formalization.} Let $s_t$ be the state of the computer at step $t$, $m_t$ the agent's working state, and $o_t$ the observation returned by the previous step, with $m_0$ containing the task instruction. At each step, a GCUA policy $\pi_\theta$ makes the three decisions jointly:
\begin{equation}
  (q_t,\ p_t,\ m_{t+1}) \sim \pi_\theta(\cdot \mid m_t,\ o_t).
  \label{eq:gcua}
\end{equation}
\emph{What to observe:} CLI/API output is returned automatically. The model selects additional views through $q_t \subseteq \mathcal{Q} = \{\text{screenshot}, \text{AX tree}\}$, where $q_t = \emptyset$ means that no additional view is requested.
\emph{How to act:} $p_t$ is a program whose statements call primitives in $\mathcal{A} = \mathcal{A}_{\mathrm{GUI}} \cup \mathcal{A}_{\mathrm{AX}} \cup \mathcal{A}_{\mathrm{CLI}} \cup \mathcal{A}_{\mathrm{API}}$ and may use loops, branches, and variables. A single primitive call is the simplest program.
\emph{How to organize its work:} $m_{t+1}$ is the updated working state, including the plan, the record of progress, and the observations kept in context.
Executing $p_t$ moves the computer to $s_{t+1}$ and produces an execution output $e_t$, such as command output or API responses. The next observation is
\begin{equation}
  o_{t+1} = \bigl(e_t,\ \{v_q(s_{t+1})\}_{q \in q_t}\bigr),
  \label{eq:obs}
\end{equation}
where $v_q$ renders view $q$ of the computer state.
The model therefore observes both the views it requests through $q_t$, such as a screenshot or an accessibility tree, and what its program prints in $e_t$.

\paragraph{A Special Case: GUI+CLI Agents.}
Our recent GUI+CLI agent, \qwenui{} \citep{qwenuiagent}, released shortly before GPT-6 Astra, is an instructive special case of Eq.~(\ref{eq:gcua}). It places GUI and CLI actions in one action space and lets the model emit a batch of several actions in one call. In our notation, it fixes
\begin{equation}
  q_t \equiv \{\text{screenshot}\}, \qquad
  p_t = (a_1, \dots, a_n),\ a_i \in \mathcal{A}_{\mathrm{GUI}} \cup \mathcal{A}_{\mathrm{CLI}}, \qquad
  m_{t+1} = h(m_t, o_t),
  \label{eq:guicli}
\end{equation}
where $h$ is a fixed harness rule, such as keeping the last $k$ screenshots. The model already chooses, action by action, whether to click or to run a command, and this alone improved both success and efficiency over GUI-only execution.
GCUA lifts each of these three constraints. Where the GUI+CLI agent emits a fixed sequence $(a_1, \dots, a_n)$, a GCUA writes a program over the full set $\mathcal{A}$, which adds accessibility and API operations as well as loops, branches, and variables. Inside such a program, a click, an accessibility operation, a shell command, and an API call are simply four kinds of statements. Where the GUI+CLI agent receives a screenshot after every step, a GCUA decides which views to request, including none at all. And where the GUI+CLI agent follows the harness rule $h$, a GCUA decides for itself what to keep in its working state.
In our formulation, batched GUI+CLI hybrid agents are a restricted case of GCUA.

\subsection{Relation to Existing Agents}
\label{sec:gcua-pos}
The formalization in Eq.~(\ref{eq:gcua}) also lets us compare GCUA with existing agents. Each existing family can be written in the same form, except that some of the three decisions (what to observe, how to act, and how to organize) are fixed in advance instead of being made by the model at each step. Table~\ref{tab:taxonomy} lists what each family fixes, and the paragraphs below discuss each family in turn.

\begin{table}[t]
\centering
\small
\caption{Existing agent families and GCUA compared on the three decisions. Gray text gives the formal notation. The observation after each step contains the views selected by $q_t$ and the execution output $e_t$ (Eq.~\ref{eq:obs}). Existing families fix some of these decisions in advance. The shaded rows mark the path this report follows, from our earlier GUI+CLI agent to GCUA, in which the model makes all three decisions at each step.}
\label{tab:taxonomy}
\vspace{0.3em}
\renewcommand{\arraystretch}{1.15}
\setlength{\tabcolsep}{4pt}
\begin{tabular}{>{\raggedright\arraybackslash}p{0.14\linewidth}
                >{\raggedright\arraybackslash}p{0.20\linewidth}
                >{\raggedright\arraybackslash}p{0.19\linewidth}
                >{\raggedright\arraybackslash}p{0.21\linewidth}
                >{\raggedright\arraybackslash}p{0.17\linewidth}}
\toprule
 & \multicolumn{3}{c}{\textbf{Decision at each step}} & \\
\cmidrule(lr){2-4}
\textbf{Family} & \textbf{What to observe} & \textbf{How to act} & \textbf{How to organize} & \textbf{Interface choice} \\
\midrule
GUI agent
 & Screenshot only\newline{\scriptsize\textcolor{gray}{$q_t \equiv \{\mathrm{screenshot}\}$}}
 & One GUI action\newline{\scriptsize\textcolor{gray}{$p_t \in \mathcal{A}_{\mathrm{GUI}}$}}
 & Harness rule, e.g., last $k$ screenshots\newline{\scriptsize\textcolor{gray}{$m_{t+1} = h(m_t, o_t)$}}
 & No choice (GUI only) \\
\addlinespace
CLI agent
 & Command output only\newline{\scriptsize\textcolor{gray}{$q_t \equiv \emptyset$, only $e_t$}}
 & Shell commands and scripts\newline{\scriptsize\textcolor{gray}{program over $\mathcal{A}_{\mathrm{CLI}}$}}
 & Partly by the model, e.g., context summaries
 & No choice (shell only) \\
\addlinespace
Multi-agent\newline hybrid
 & Fixed for each sub-agent\newline{\scriptsize\textcolor{gray}{$q_t \equiv \mathcal{Q}_k$}}
 & Fixed for each sub-agent\newline{\scriptsize\textcolor{gray}{$p_t$ over $\mathcal{A}_k$}}
 & Split between orchestrator and sub-agents
 & Orchestrator, per subtask \\
\midrule
\rowcolor[gray]{0.975}
\qwenui{}\newline(GUI+CLI)
 & Screenshot every step, plus command output when a command runs\newline{\scriptsize\textcolor{gray}{$q_t \equiv \{\mathrm{screenshot}\}$, plus $e_t$}}
 & A batch of GUI and CLI actions\newline{\scriptsize\textcolor{gray}{sequence over $\mathcal{A}_{\mathrm{GUI}} \cup \mathcal{A}_{\mathrm{CLI}}$}}
 & Harness rule\newline{\scriptsize\textcolor{gray}{$m_{t+1} = h(m_t, o_t)$}}
 & Model, per action (GUI or shell) \\
\rowcolor[gray]{0.94}
GCUA
 & Chosen per step, plus program output\newline{\scriptsize\textcolor{gray}{$q_t \subseteq \mathcal{Q}$, plus $e_t$}}
 & Program over all interfaces\newline{\scriptsize\textcolor{gray}{program over $\mathcal{A}$}}
 & Decided by the model\newline{\scriptsize\textcolor{gray}{$m_{t+1} \sim \pi_\theta$}}
 & Model, per step \\
\bottomrule
\end{tabular}
\end{table}

\paragraph{GUI Agents.} A GUI agent fixes all three decisions \citep{uitars, opencua, lu2026qwen}. It always observes a screenshot, so $q_t \equiv \{\text{screenshot}\}$. It acts through one GUI operation per step, so $p_t$ is a single primitive in $\mathcal{A}_{\mathrm{GUI}}$. Its working state follows a harness rule $m_{t+1} = h(m_t, o_t)$, such as keeping the last $k$ screenshots. The model still chooses where to click, but not what to observe, which interface to use, or what to keep.

\paragraph{CLI Agents.} A CLI agent fixes what it observes and restricts how it acts. It has no view of the screen, so $q_t \equiv \emptyset$ and it sees only the execution output $e_t$. It acts only through the shell, so $p_t$ is a program over $\mathcal{A}_{\mathrm{CLI}}$. It is more flexible than a GUI agent. Its commands and scripts can use loops, branches, and variables. Recent coding agents also let the model decide part of how it organizes its work, for example by summarizing earlier context when it grows long \citep{claudecode, codexcli}. What a CLI agent cannot do is operate an application that can only be used through its graphical interface. Agent frameworks such as OpenClaw support more than the shell, but they also come without built-in GUI control \citep{agentslastexam}.

\paragraph{Multi-Agent Hybrid Systems.} Some recent systems combine GUI and programmatic control by dividing the work among several agents. For example, an orchestrator splits a task into subtasks and sends each one to either a GUI operator or a programmer \citep{coact1}. Each sub-agent has its own fixed interface. The GUI operator sees screenshots and acts through GUI operations, and the programmer writes and runs code. In terms of Eq.~(\ref{eq:gcua}), while sub-agent $k$ handles a subtask, $q_t$ and $p_t$ are limited to its own sets $\mathcal{Q}_k$ and $\mathcal{A}_k$. Together, the sub-agents cover the same interfaces as a GCUA. The difference is where the choice of interface is made. In these systems, the orchestrator makes it once per subtask. In a GCUA, a single model makes it at every step, and one program can combine several interfaces.

\subsection{Why Now: The Route Flip to Code Execution}
\label{sec:gcua-why}
In 2026, the frontier began to shift from clicking toward programming, and the shift is visible in vendor documentation rather than only in benchmark scores.
With the release of GPT-6 Astra, OpenAI's computer-use guide began recommending that models write and run short programs, for example with PyAutoGUI or Playwright, rather than rely primarily on the earlier computer tool for mouse and keyboard actions \citep{openai2026cuaguide}. The guide's examples reflect this shift: the computer tool still targets the previous model (i.e., GPT-5.6 Sol), while code-execution examples use GPT-6 Astra. A similar pattern appears on \osworldtwo{}, where frontier agents perform more actions through code, APIs, and file operations rather than clicks \citep{osworld2}.
OpenAI also reports that GPT-6 Astra completes OSWorld 2.0 tasks in about 47\% less time than GPT-5.6 Sol and is 1.9$\times$ faster on Mind2Web \citep{openai2026gpt6astra}.
OpenAI does not say how much of this gain comes from code execution, but it is the kind of gain one would expect when each model call does more work.

\textbf{Code Execution versus Batched Actions.} With batched actions, each model call yields a list of actions that were fixed when the call was generated: the model must decide every click before it sees the result of the first one. With code execution, each model call yields a program, and control flow is decided while the program runs. Three things become possible in a single call that were impossible before: a \emph{condition} (\texttt{if the dialog is visible, dismiss it}), a \emph{loop} (\texttt{for each row, fill the three fields}), and a \emph{verification} (\texttt{assert the status now reads Saved}). A task that took a dozen round trips becomes one. This is also why batched actions, which our earlier work introduced \citep{qwenuiagent}, are a stepping stone rather than an endpoint: a batch is a program without control flow.

\subsection{Efficiency as a First-Class Metric}
\label{sec:gcua-eff}
Success rate remains our primary measure, but it does not capture the resources required to complete a task.
For interactive tasks, an important threshold is reached when delegation delivers results sooner than manual execution, potentially saving both human effort and time to completion.
We therefore evaluate efficiency using four metrics: steps per task, output tokens, reasoning tokens, and cost.

Each step is one model call, which may generate multiple actions through batching or code execution.
Fewer steps reduce model-call overhead, while fewer output tokens can lower generation time and cost.
Reasoning tokens are included in output tokens but reported separately.
When reasoning accounts for a large share of output, reducing unnecessary reasoning can improve efficiency without shortening the executable code, provided task performance is maintained.
We compute cost from input and output token usage at each model's published prices, including cache discounts.

\section{Agent Design: Code Execution and Harness-to-Model Decision Transfer}
\label{sec:model}
\vspace{-0.5em}
This section studies two behaviors we aim to develop in GCUA models. The first is \textbf{code execution}, which concerns how to act: the model combines graphical interaction and programmatic operations within a single program. The second is \textbf{harness-to-model decision transfer}, which concerns what to observe and how to organize: the model makes selected workflow control decisions that a harness would otherwise fix with rules, such as when to look at the screen and how to organize task state. This release does not introduce a trained model. Instead, we study how the prompts, tool interfaces, and execution runtime of \projname{} support these behaviors in existing frontier models. We first explain the design and illustrate it with execution trajectories, then evaluate its effects on task completion and cost.

\subsection{Code Execution over GUI and CLI}
\label{sec:code-exec}

Code execution gives graphical interaction and programmatic operations a unified executable form. The model writes a program in which GUI actions, such as \texttt{pyautogui.click()}, can be combined with shell commands and Python computation. The model can therefore specify not only what operations to perform, but also how execution should proceed. Loops and conditional logic allow the program to repeat operations and choose subsequent actions based on intermediate results, without requiring a new model call for each operation. On \osworldtwo{}, this design improves \qwenmax{}'s mean task score by 5.2 points and reduces steps per task by 17.0\% relative to a hybrid CUA baseline with batched GUI+CLI actions (Section~\ref{sec:model-results}).

\subsubsection{Implementation}
\textbf{Prompt Design.}
The system prompt specifies the programming interfaces and output format for code execution. The OSWorld-v2 configuration specifies PyAutoGUI for desktop interaction and the Python function \texttt{bash()} for executing shell commands.
The prompt for the real-device harness additionally describes Playwright for browser interaction. The prompt requires one Python program in a \texttt{<code>} block per step and includes examples that combine interface operations, shell commands, and checks on intermediate results. The full prompt is given in Appendix \ref{app:prompts}.

\textbf{Program Execution.}
The harness extracts and syntax-checks the code block, then executes it in the computer environment. The runtime supports desktop interaction through PyAutoGUI and, in the real-device harness, browser interaction through Playwright.
Shell commands execute through \texttt{bash()}, which returns command output and exit status. The Python control flow coordinates these calls, allowing loops and conditions to combine GUI and shell operations within one program.

\textbf{Execution Feedback.}
The harness returns captured standard output, standard error, and execution status to the model. With on-demand observation enabled, observations include a new screenshot when the program requests one through the \texttt{screenshot()} action. The real-device harness additionally returns accessibility-tree observations when requested by the model.
We have open-sourced both the preliminary real-device harness and the code execution harness.

\subsubsection{Representative Code-Execution Patterns}
\label{sec:code-patterns}

The implementation specifies what a program can contain, but we want to see how models actually write programs to solve concrete tasks.
We therefore examine GPT-6 Astra on \osworldtwo{}, and find five recurring patterns in how its programs connect actions, intermediate results, and model observations. The examples illustrate complementary uses of code, rather than mutually exclusive categories or isolated performance gains.

\begin{figure}[!t]
    \centering
    \includegraphics[width=\linewidth]{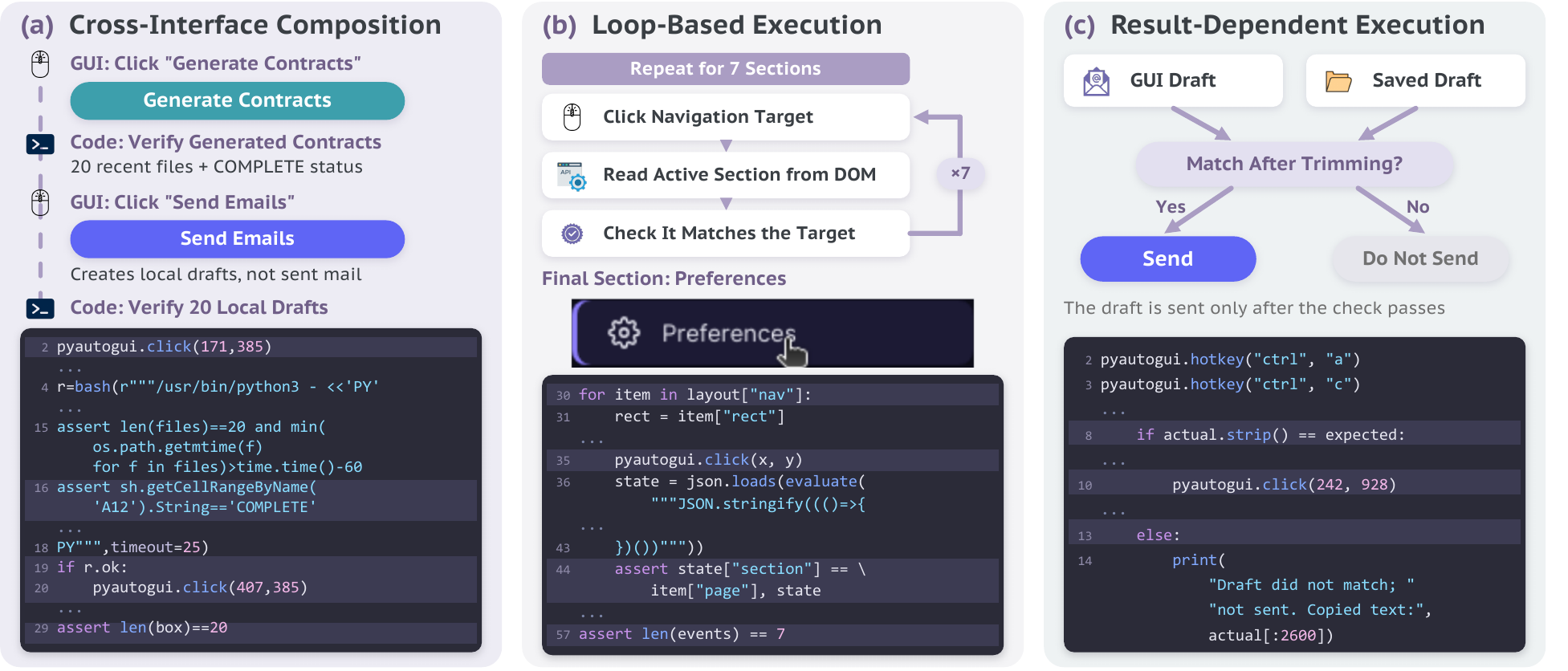}
    \caption{\textbf{Programs coordinate actions and checks.}
    (a) GUI clicks alternate with checks of generated contracts and local drafts.
    (b) A loop visits seven webpage sections and checks each active section through the DOM.
    (c) A draft comparison determines whether to click Send, without verifying delivery.}
    \label{fig:patterns-execution}
\end{figure}

\begin{figure}[!t]
    \centering
    \includegraphics[width=\linewidth]{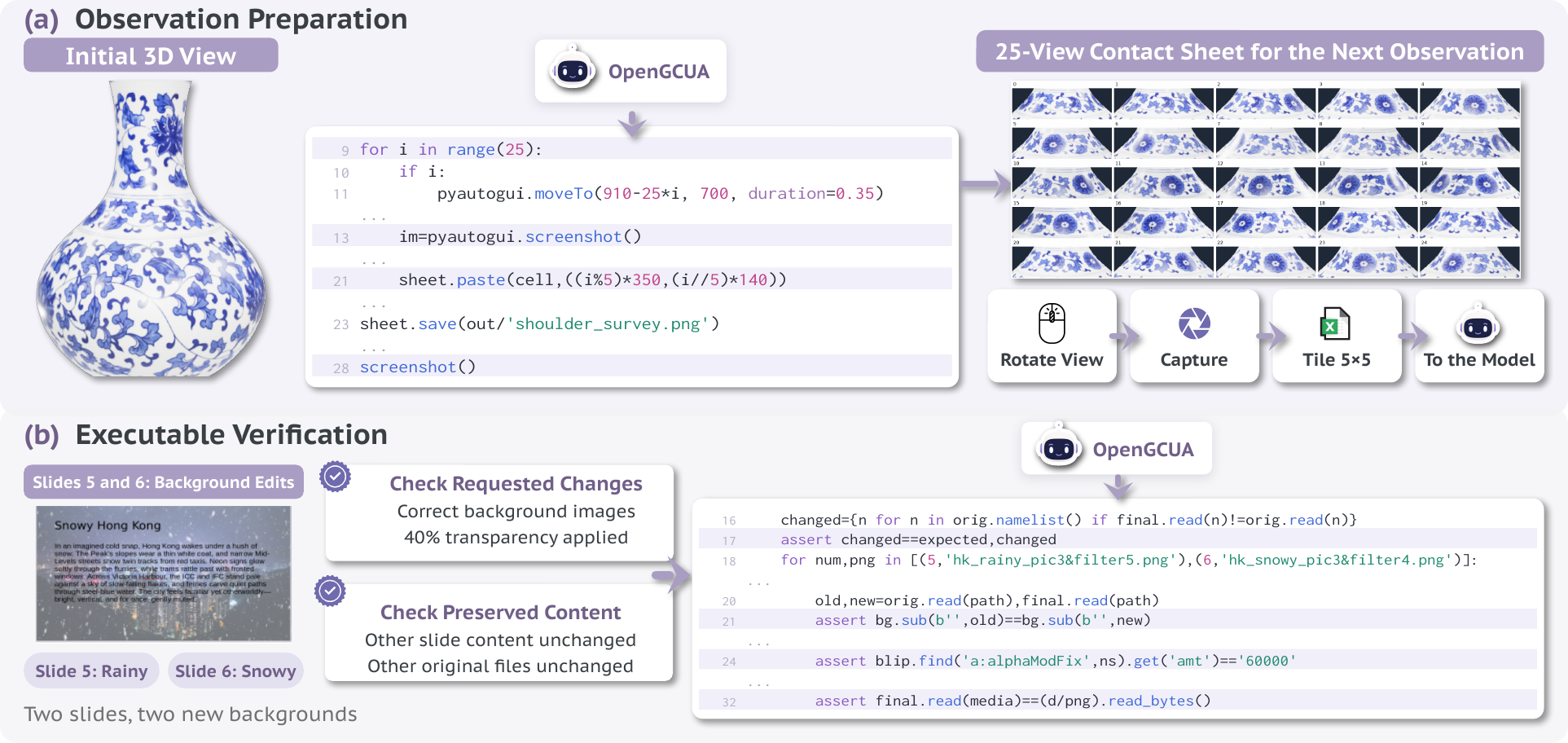}
    \caption{\textbf{Programs prepare and verify evidence.}
    (a) A program assembles 25 views into a contact sheet for the model's next observation.
    (b) Assertions check background images and transparency, and verify that non-background content remains unchanged.}
    \label{fig:patterns-evidence}
\end{figure}

\textbf{Cross-Interface Composition.}
A single program can use information from one interface to guide operations through another. 
A program can alternate between GUI actions and programmatic checks within one model step. In a contract-generation task, the agent clicks a spreadsheet button, reads the application's completion status, and checks the output files. Only after these checks pass does it click a second button to create local email drafts, which it then inspects programmatically. This example connects GUI execution with structured feedback from both the application and its files (Figure~\ref{fig:patterns-execution}a).

\textbf{Loop-Based Execution.}
Loops let the model specify a procedure once and repeat it across multiple targets within one program. Unlike a fixed action list, the targets and number of iterations can be determined during execution, for example from file contents or application state. This allows repeated work to proceed without generating a separate action sequence or making a new model call for each iteration. 
To record a website walkthrough, the agent iterates over seven navigation targets, clicks each target, and reads the active section from the DOM. An assertion checks that the resulting section matches the intended target before the loop continues. The program handles all seven visits and checks without returning to the model between iterations (Figure~\ref{fig:patterns-execution}b).

\textbf{Result-Dependent Execution.}
Conditional logic lets the program respond to what happens during execution. It can inspect command output or application state, check a result, and decide whether to continue. Such decisions can remain within the program when their conditions are computable, rather than requiring the model to interpret each intermediate result. A local check can determine whether a GUI action should proceed. Before sending an email reply, the agent copies the draft from the interface and compares it with a saved version after trimming leading and trailing whitespace. Its program clicks Send only if the texts match and reports a mismatch otherwise. The recorded execution follows the matching branch, but delivery is not verified within this step (Figure~\ref{fig:patterns-execution}c).

\textbf{Observation Preparation.}
Programs can gather and organize evidence for the next model decision. They can collect multiple views, crop relevant regions, or arrange images into a visual summary. This extends code execution from performing actions to preparing observations that the model can then interpret. While inspecting a three-dimensional object, the agent collects 25 views, crops the relevant region from each, and assembles a numbered contact sheet. It opens this sheet and requests one new visual observation. The program collects and arranges views for the model's visual judgment (Figure~\ref{fig:patterns-evidence}a).

\textbf{Executable Verification.}
Programs can check both intended changes and content that should remain unchanged. After editing two slide backgrounds, the agent compares the original and saved presentation packages. It checks the expected changes, verifies that the other 64 original package entries are unchanged, and compares slide content outside the background elements. These assertions make selected preservation requirements explicit, but do not replace independent evaluation (Figure~\ref{fig:patterns-evidence}b).

These examples show a useful division of work: programs execute locally specified operations and checks, while the model chooses the procedure and interprets evidence that still requires judgment.

\subsubsection{Limits of Visual-Only Feedback}
\label{sec:code-limits}

Result-dependent control is harder when interface observations are limited to screenshots. Conditions such as whether a button is enabled or an input field contains the expected value are not directly exposed as program-readable values, so checking them may require another model call. Accessibility trees can expose text and control states as structured data, allowing programs to query supported interface properties and use them in loops and conditional logic. This motivates combining visual observation with accessibility-tree access in our real-device harness design.

\subsection{Harness-to-Model Decision Transfer}
\label{sec:selfdirected}
Code execution addresses \emph{how to act} by allowing the model to express computer operations as programs. A GCUA must also decide \emph{what to observe} and \emph{how to organize} its work. We study \emph{harness-to-model decision transfer}: replacing selected fixed harness rules with task-dependent model decisions. A screenshot, for example, may be necessary after a GUI action but redundant when program output already answers the relevant question. Deciding whether to request it should depend on the information needed to continue.

\textbf{Mechanism in the Harness, Decision in the Model.}
As models become more capable, we envision progressively shifting selected workflow decisions from fixed harness rules to the model. The harness should provide reliable mechanisms, while the model decides when and how to use them based on the task and available evidence. In the current implementation, screenshot requests and task-state updates are model-initiated workflow operations, complemented by model-generated action-intent annotations. The harness continues to execute interface calls, preserve task records, and manage context retention.

\subsubsection{Mechanisms for Harness-to-Model Decision Transfer}

\textbf{On-Demand Observation.}
The \texttt{screenshot} function lets the model decide whether its next input needs a new screenshot. It can request visual feedback after a GUI action or continue without it when text output is sufficient. On-demand observation can reduce context length by omitting screenshots that add little information. The environment still captures a screenshot after every step, so the mechanism changes what the model receives, not what the environment records.

\textbf{Task-State Tracking.}
The \texttt{todo} interface lets the model maintain a persistent record of progress, facts, and unresolved questions. It updates this record through \texttt{todo.update()} and retrieves it through \texttt{todo.list()}. 
Statuses distinguish pending work, attempted changes, unverified changes, and claimed verification, helping the model track what still needs checking. These entries record the model's assessment rather than independently verifying the environment.

\textbf{Action Intent.}
The \texttt{intent} annotation makes the purpose of each code-execution step explicit. Before each program runs, the model provides a short \texttt{<intent>} block stating what the code is intended to accomplish. A program may combine many operations, making its overall purpose difficult to infer from the code alone. 
In real-device use, the harness can display this description to users, helping them understand what the agent is trying to do at each step without reading the program.

\textbf{Harness-Managed Context.}
The harness manages historical tool outputs and screenshots through two fixed policies. A \texttt{rolling screenshot buffer} removes older images in batches, limiting visual context while reducing disruption to prompt caching \citep{anthropic2026computerusepractices}. \texttt{Micro-compaction} applies a similar rolling-buffer policy to textual tool outputs, replacing older outputs with short placeholders in batches while retaining recent results. These retention decisions remain with the harness, complementing the model's choices about when to obtain new observations and how to track task progress.

\subsubsection{Representative Harness-to-Model Decision Transfer Patterns}
\label{sec:workflow-patterns}
\begin{figure}[t]
    \centering
    \includegraphics[width=\linewidth]{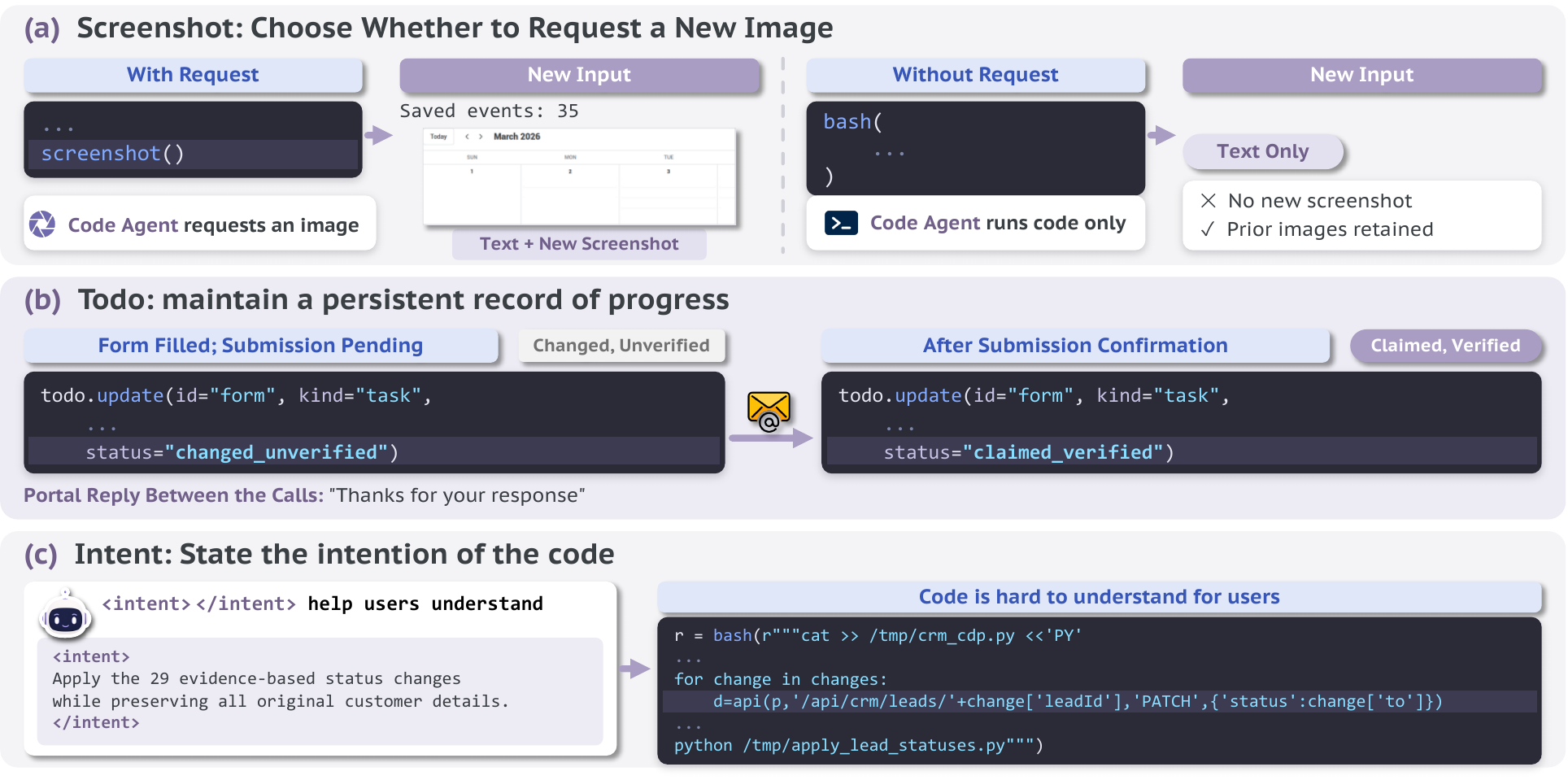}
    \caption{\textbf{Model outputs for observation, task state, and intent.}
    (a) The two programs differ in whether they request a new screenshot for the next input.
    (b) A form record remains unverified until the model reads submission confirmation.
    (c) Intent states what the program should change and preserve.}
    \label{fig:workflow-patterns}
\end{figure}

Figure~\ref{fig:workflow-patterns} shows these interfaces in model outputs. The examples come from GPT-6 Astra trajectories on \osworldtwo{}. They illustrate the mechanisms, not their individual effects on task success.

\textbf{Requesting Visual Feedback Only When Needed.}
In a calendar task, the model checks saved events and calls \texttt{screenshot()}. Its next input contains text output and a new screenshot. The following program inspects event styles without requesting another image, so the next input contains text feedback but no new screenshot. These adjacent steps show the observation decision: the model can request visual input or continue with programmatic feedback. Earlier screenshots may still remain in context.

\textbf{Keeping Unfinished Work Explicit across Programs.}
In a form-filling task, the model records that all fields have been entered but submission is still pending. It marks the task \texttt{changed\_unverified}. After a later program clicks Submit and reads the page's confirmation, the model updates the same record to \texttt{claimed\_verified}. The persistent record distinguishes work performed from an outcome checked. Its status remains the model's assessment, not an independent evaluation.

\textbf{Stating a Program's Purpose before Execution.}
In a CRM task, one intent statement summarizes a 64-line program: update 29 lead statuses while preserving the original customer details. The program performs status-only updates and checks other fields, allowing modification timestamps to change. The \texttt{<intent>} block makes both the intended change and the preservation requirement explicit without requiring the reader to reconstruct them from the code.

\subsection{Experiments on Agent Design}
\label{sec:model-results}
We evaluate whether the agent design presented in this report helps frontier models complete computer tasks more effectively. We first compare the full GCUA configuration with a GUI-only baseline that already supports batched actions. We then separate the effects of interface access, code execution, and harness-to-model decision transfer.

\subsubsection{Models and Configuration}
\textbf{Benchmark and Models.}
Following the evaluation protocol of the GPT-6 Astra release \citep{openai2026gpt6astra}, we evaluate on the 82-task offline subset of \osworldtwo{} (v2.1) \citep{osworld2}, which excludes the 26 tasks that require internet access and so keeps results independent of external websites at run time. The main comparison covers eight models: Claude Opus 5, GPT-6 Astra, Gemini 3.8 Flash, \qwenmax{}, \qwenflash{}, Kimi K3, DeepSeek-V4.1-Flash, and GLM-5.3-Flash. 

\textbf{Execution Configurations.}
The CUA baseline uses GUI-only interaction with batched actions enabled. The full Open Generalist CUA (\textbf{OpenGCUA}) configuration combines code execution over GUI and shell operations with the model-initiated workflow control and harness-managed context policies described above. For the ablations of code execution, harness-to-model decision transfer, and harness-managed context policies, we explicitly specify which components are enabled in each setting.

\textbf{Metrics.}
We evaluate task performance and execution efficiency. Following \osworldtwo{}, we report binary success and partial-credit scores to measure full and partial task completion, respectively. We measure efficiency by mean steps per task and mean output tokens per task.

\subsubsection{CUA versus GCUA}

\begin{table}[!t]
\centering
\small
\setlength{\tabcolsep}{4pt}
\renewcommand{\arraystretch}{1.20}
\caption{\textbf{CUA versus OpenGCUA on the OSWorld-v2 offline set.} Each model is evaluated with a GUI-only baseline that supports batched actions and the full GCUA configuration with code execution and harness-to-model decision transfer. Bold marks the better value within each model and metric. K denotes 1,000 tokens. }
\label{tab:main}
\begin{tabular*}{\linewidth}{@{\extracolsep{\fill}}llrrrr@{}}
\toprule[1.1pt]
\textbf{Model} & \textbf{Configuration} & \multicolumn{2}{c}{\textbf{Task completion}} & \multicolumn{2}{c}{\textbf{Interaction cost}} \\
\cmidrule(lr){3-4}\cmidrule(lr){5-6}
 &  & Score $\uparrow$ & Success $\uparrow$ & Steps/task $\downarrow$ & Output K/task $\downarrow$ \\
\midrule
\multirow{2}{*}{Claude Opus 5} & GUI + batch & \cellcolor{baselineScore}\makebox[2.2em][r]{60.7}\makebox[2.8em][r]{\scriptsize } & 29.3 & \cellcolor{baselineScore}\makebox[2.7em][r]{103.1}\makebox[3.7em][r]{\scriptsize } & 378.3 \\
 & \textbf{OpenGCUA} & \cellcolor{ogcuaScore}\makebox[2.2em][r]{\textbf{70.1}}\makebox[2.8em][r]{\scriptsize (+9.4)} & \textbf{36.6} & \cellcolor{ogcuaScore}\makebox[2.7em][r]{\textbf{61.4}}\makebox[3.7em][r]{\scriptsize ($-40.4\%$)} & \textbf{89.3} \\
\addlinespace[0.25em]
\multirow{2}{*}{GPT-6 Astra} & GUI + batch & \cellcolor{baselineScore}\makebox[2.2em][r]{67.1}\makebox[2.8em][r]{\scriptsize } & 36.6 & \cellcolor{baselineScore}\makebox[2.7em][r]{102.1}\makebox[3.7em][r]{\scriptsize } & \textbf{30.8} \\
 & \textbf{OpenGCUA} & \cellcolor{ogcuaScore}\makebox[2.2em][r]{\textbf{75.7}}\makebox[2.8em][r]{\scriptsize (+8.7)} & \textbf{46.3} & \cellcolor{ogcuaScore}\makebox[2.7em][r]{\textbf{33.5}}\makebox[3.7em][r]{\scriptsize ($-67.2\%$)} & 31.9 \\
\addlinespace[0.25em]
\multirow{2}{*}{Gemini 3.8 Flash} & GUI + batch & \cellcolor{baselineScore}\makebox[2.2em][r]{50.1}\makebox[2.8em][r]{\scriptsize } & 22.0 & \cellcolor{baselineScore}\makebox[2.7em][r]{166.8}\makebox[3.7em][r]{\scriptsize } & \textbf{80.3} \\
 & \textbf{OpenGCUA} & \cellcolor{ogcuaScore}\makebox[2.2em][r]{\textbf{67.0}}\makebox[2.8em][r]{\scriptsize (+16.9)} & \textbf{31.7} & \cellcolor{ogcuaScore}\makebox[2.7em][r]{\textbf{101.2}}\makebox[3.7em][r]{\scriptsize ($-39.3\%$)} & 82.0 \\
\addlinespace[0.25em]
\multirow{2}{*}{Qwen3.8-Max} & GUI + batch & \cellcolor{baselineScore}\makebox[2.2em][r]{46.9}\makebox[2.8em][r]{\scriptsize } & 17.1 & \cellcolor{baselineScore}\makebox[2.7em][r]{127.7}\makebox[3.7em][r]{\scriptsize } & 122.3 \\
 & \textbf{OpenGCUA} & \cellcolor{ogcuaScore}\makebox[2.2em][r]{\textbf{64.0}}\makebox[2.8em][r]{\scriptsize (+17.1)} & \textbf{28.9} & \cellcolor{ogcuaScore}\makebox[2.7em][r]{\textbf{57.0}}\makebox[3.7em][r]{\scriptsize ($-55.4\%$)} & \textbf{99.5} \\
\addlinespace[0.25em]
\multirow{2}{*}{Qwen3.8-Flash} & GUI + batch & \cellcolor{baselineScore}\makebox[2.2em][r]{38.9}\makebox[2.8em][r]{\scriptsize } & 14.2 & \cellcolor{baselineScore}\makebox[2.7em][r]{232.0}\makebox[3.7em][r]{\scriptsize } & 189.8 \\
 & \textbf{OpenGCUA} & \cellcolor{ogcuaScore}\makebox[2.2em][r]{\textbf{60.5}}\makebox[2.8em][r]{\scriptsize (+21.6)} & \textbf{27.6} & \cellcolor{ogcuaScore}\makebox[2.7em][r]{\textbf{86.2}}\makebox[3.7em][r]{\scriptsize ($-62.8\%$)} & \textbf{159.6} \\
\addlinespace[0.25em]
\multirow{2}{*}{Kimi K3} & GUI + batch & \cellcolor{baselineScore}\makebox[2.2em][r]{40.7}\makebox[2.8em][r]{\scriptsize } & 15.9 & \cellcolor{baselineScore}\makebox[2.7em][r]{160.1}\makebox[3.7em][r]{\scriptsize } & 302.5 \\
 & \textbf{OpenGCUA} & \cellcolor{ogcuaScore}\makebox[2.2em][r]{\textbf{59.2}}\makebox[2.8em][r]{\scriptsize (+18.5)} & \textbf{22.0} & \cellcolor{ogcuaScore}\makebox[2.7em][r]{\textbf{81.9}}\makebox[3.7em][r]{\scriptsize ($-48.8\%$)} & \textbf{139.6} \\
\addlinespace[0.25em]
\multirow{2}{*}{DeepSeek-V4.1-Flash} & GUI + batch & \cellcolor{baselineScore}\makebox[2.2em][r]{28.0}\makebox[2.8em][r]{\scriptsize } & 8.5 & \cellcolor{baselineScore}\makebox[2.7em][r]{167.3}\makebox[3.7em][r]{\scriptsize } & 316.0 \\
 & \textbf{OpenGCUA} & \cellcolor{ogcuaScore}\makebox[2.2em][r]{\textbf{57.7}}\makebox[2.8em][r]{\scriptsize (+29.8)} & \textbf{26.8} & \cellcolor{ogcuaScore}\makebox[2.7em][r]{\textbf{84.9}}\makebox[3.7em][r]{\scriptsize ($-49.3\%$)} & \textbf{180.4} \\
\addlinespace[0.25em]
\multirow{2}{*}{GLM-5.3-Flash} & GUI + batch & \cellcolor{baselineScore}\makebox[2.2em][r]{29.6}\makebox[2.8em][r]{\scriptsize } & 14.6 & \cellcolor{baselineScore}\makebox[2.7em][r]{190.8}\makebox[3.7em][r]{\scriptsize } & 155.7 \\
 & \textbf{OpenGCUA} & \cellcolor{ogcuaScore}\makebox[2.2em][r]{\textbf{64.9}}\makebox[2.8em][r]{\scriptsize (+35.2)} & \textbf{29.3} & \cellcolor{ogcuaScore}\makebox[2.7em][r]{\textbf{75.5}}\makebox[3.7em][r]{\scriptsize ($-60.4\%$)} & \textbf{131.0} \\
\midrule
\multicolumn{6}{c}{\textbf{Mean gain: +19.6 score points \quad 53.0\% fewer steps \quad 27.2\% fewer output tokens}}\\
\bottomrule[1.1pt]
\end{tabular*}
\end{table}

We compare the conventional GUI-only CUA setting with the full OpenGCUA configuration to examine the performance and efficiency gains from moving toward generalist computer use. OpenGCUA gives the model greater control over what to observe, how to act, and how to organize its work. The CUA baseline already supports batched actions, while OpenGCUA combines code execution with harness-to-model decision transfer.

\textbf{OpenGCUA improves task scores and reduces steps across all eight models.} As shown in Table~\ref{tab:main}, the cross-model mean partial score increases from 45.3 to 64.9, a gain of \textbf{19.6 points}, while steps per task decrease by an average of \textbf{53.0\%}, computed from each model's relative reduction. The largest score gains reach \textbf{35.2 points} for GLM-5.3-Flash and \textbf{29.8 points} for DeepSeek-V4.1-Flash. Full-task success also increases for every model. These results show a substantial gap between batched GUI-only execution and the full OpenGCUA design in both task performance and interaction efficiency.

\textbf{The gains hold despite large differences in baseline performance and efficiency.} The benefits are not limited to models with low initial task scores. GPT-6 Astra, which has the highest baseline score, improves from 67.1 to 75.7 and reduces steps per task from 102.1 to 33.5, a \textbf{67.2\%} reduction. At the other end of the efficiency range, \qwenflash{} starts with 232.0 steps per task, yet gains \textbf{21.6 points} while using \textbf{62.8\%} fewer steps. The consistent improvements across all 8 frontier models support the broad applicability of the OpenGCUA design.

\textbf{Fewer interaction steps do not necessarily require fewer output tokens.}
Output tokens per task decrease for six of the eight models, with an average relative reduction of \textbf{27.2\%}. Claude Opus 5 reduces its output from 378.3K to 89.3K tokens per task, and Kimi K3 from 302.5K to 139.6K. GPT-6 Astra, however, uses approximately the same output-token budget, despite its large reduction in steps. Fewer steps per task can reduce model round trips and their overhead, even without reducing output tokens. End-to-end latency must be measured separately, accounting for model response time and tool execution time. We report both steps and output tokens to distinguish interaction efficiency from output-token usage.

\begin{figure}[t]
    \centering
    \includegraphics[width=\linewidth]{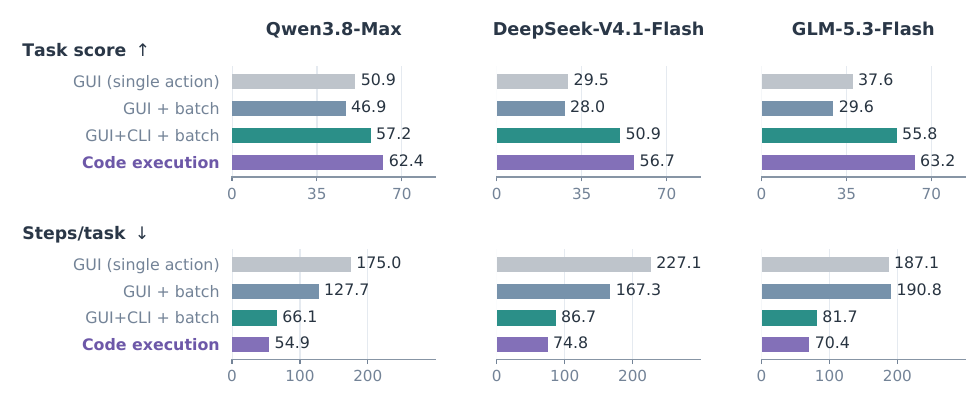}
    \caption{\textbf{Code-execution ablation on OSWorld-v2.}
    Code execution achieves higher task scores and fewer steps than single-action GUI, batched GUI, and batched GUI+CLI setting across all three models.}
    \label{fig:codeexec-ablation}
\end{figure}
\subsubsection{GUI+CLI Hybrid CUA versus GCUA}

\begin{wraptable}[8]{r}{0.52\textwidth}
\vspace{-12pt}  
\centering
\fontsize{8.3}{10.2}\selectfont
\setlength{\tabcolsep}{1.8pt}
\renewcommand{\arraystretch}{1.12}
\captionsetup{font=small,justification=raggedright,singlelinecheck=false,skip=5pt}
\caption{\textbf{Full GCUA versus batched GUI+CLI.}}
\label{tab:cua-gcua}
\begin{tabular*}{\linewidth}{@{\extracolsep{\fill}}l rr rr@{}}
\toprule[1.1pt]
Model & \multicolumn{2}{c}{Task score $\uparrow$} & \multicolumn{2}{c}{Steps/task $\downarrow$} \\
\cmidrule(lr){2-3}\cmidrule(lr){4-5}
 & GUI+CLI & GCUA & GUI+CLI & GCUA \\
\midrule
Qwen3.8-Max & \cellcolor{baselineScore}57.2 & \cellcolor{ogcuaScore}64.0 & \cellcolor{baselineScore}66.1 & \cellcolor{ogcuaScore}57.0 \\
DeepSeek-V4.1-Flash & \cellcolor{baselineScore}50.9 & \cellcolor{ogcuaScore}57.7 & \cellcolor{baselineScore}86.7 & \cellcolor{ogcuaScore}84.9 \\
GLM-5.3-Flash & \cellcolor{baselineScore}55.8 & \cellcolor{ogcuaScore}64.9 & \cellcolor{baselineScore}81.7 & \cellcolor{ogcuaScore}75.5 \\
Claude Opus 5 & \cellcolor{baselineScore}65.7 & \cellcolor{ogcuaScore}70.1 & \cellcolor{baselineScore}61.6 & \cellcolor{ogcuaScore}61.4 \\
\bottomrule[1.1pt]
\end{tabular*}
\end{wraptable}

GUI+CLI hybrid agents have become an active direction in computer-use research~\citep{coact1}, including our previous work, \qwenui{}~\citep{qwenuiagent}. As discussed in Section~\ref{sec:gcua-def}, the batched GUI+CLI formulation is a special case of GCUA. We compare this baseline with the full OpenGCUA setting. Both provide GUI and CLI access, but OpenGCUA introduces code execution and model-initiated workflow control, letting the model decide what to observe, how to act, and how to organize its work.

\textbf{OpenGCUA achieves higher task scores with fewer reported steps.}
In the selected runs, OpenGCUA increases task scores by \textbf{4.4 to 9.1 points} across all four models, with lower reported steps per task (Table~\ref{tab:cua-gcua}). \qwenmax{} improves from 57.2 to 64.0 with \textbf{13.8\% fewer steps}, while GLM-5.3-Flash improves from 55.8 to 64.9 with \textbf{7.6\% fewer steps}. These results suggest that effective hybrid computer use depends not only on interface access, but also on control over execution and workflow. Code execution lets intermediate results determine subsequent GUI and shell operations within a single program. Harness-to-model decision transfer extends the model's control to when visual feedback is requested and how task progress is recorded. The comparison evaluates these mechanisms jointly.

\subsubsection{Code-Execution Ablation}
\textbf{Code execution improves task scores and reduces interaction steps.}
We compare four execution configurations for \qwenmax{}, DeepSeek-V4.1-Flash, and GLM-5.3-Flash: single-action GUI execution, batched GUI actions, batched GUI+CLI actions, and code execution over GUI and CLI operations. In the selected runs, code execution achieves higher task scores and fewer steps than each alternative across all three models (Figure~\ref{fig:codeexec-ablation}). Relative to batched GUI actions, task scores improve by \textbf{15.5 to 33.6 points}, while steps per task decrease by \textbf{55.3\% to 63.1\%}.

\textbf{Executable control adds value beyond GUI+CLI access.}
The batched GUI+CLI baseline provides the same interfaces and already supports shell scripts. Nevertheless, code execution raises GLM-5.3-Flash's task score from 55.8 to 63.2, a gain of \textbf{7.4 points}, and reduces \qwenmax{}'s steps per task from 66.1 to 54.9, a \textbf{17.0\% reduction}. The key distinction is how operations are coordinated within a step. A batch fixes the sequence of GUI and shell actions in advance, whereas a program can use intermediate results to determine subsequent operations across both interfaces. Loops, conditional branches, and retries can thus consolidate work that might otherwise require multiple model calls. This unified interface also lets models apply their existing programming capabilities directly to computer interaction, rather than expressing computer interaction through a separate action-list format. Together, these properties may help explain the observed improvements in task scores and interaction efficiency.

\subsubsection{Harness-to-Model Decision Transfer Ablation}
\textbf{Most mechanisms are not designed primarily to improve task scores.}
We ablate the mechanisms supporting harness-to-model decision transfer and the harness-managed context policies. Using code execution as the baseline, we add each feature separately and then combine all features for \qwenmax{} and DeepSeek-V4.1-Flash (Table~\ref{tab:selfdirected-abl}). While \texttt{todo} aims to support task completion, the other features primarily target observation efficiency, interpretability, or context management. We therefore assess cache reuse where relevant and check whether task scores are maintained.

\begin{table}[!t]
\centering
\small
\setlength{\tabcolsep}{2.4pt}
\renewcommand{\arraystretch}{1.16}
\caption{\textbf{Ablation of harness-to-model decision transfer and harness-managed context policies.} Each mechanism is added separately to code execution; the final row combines all mechanisms. Cache hit is the fraction of input tokens served from cache; dashes indicate omitted rates.}
\label{tab:selfdirected-abl}
\begin{tabular*}{\linewidth}{@{\extracolsep{\fill}}l rrr rrr@{}}
\toprule[0.9pt]
Configuration & \multicolumn{3}{c}{Qwen3.8-Max} & \multicolumn{3}{c}{DeepSeek-V4.1-Flash} \\
\cmidrule(lr){2-4}\cmidrule(lr){5-7}
 & Score $\uparrow$ & Steps/task $\downarrow$ & Cache hit (\%) & Score $\uparrow$ & Steps/task $\downarrow$ & Cache hit (\%) \\
\midrule
\rowcolor{baselineScore}
Code execution & 62.4 & 54.9 & 75.6 & 56.7 & 74.5 & 61.9 \\
\midrule
+ screenshot & 64.3 & 63.0 & 86.1 & 58.1 & 90.0 & 76.3 \\
+ todo & 64.0 & 56.7 & -- & 58.5 & 74.1 & -- \\
+ intent & 61.5 & 53.4 & -- & 53.3 & 82.4 & -- \\
\midrule
+ micro-compaction & 63.4 & 58.7 & 68.6 & 56.8 & 79.7 & 58.9 \\
+ rolling screenshot buffer & 64.4 & 53.2 & 91.6 & 57.2 & 79.2 & 75.3 \\
\midrule
\rowcolor{ogcuaScore}
\textbf{+ all mechanisms} & 64.0 & 57.0 & 84.1 & 57.7 & 84.9 & 70.5 \\
\bottomrule[0.9pt]
\end{tabular*}
\end{table}

\textbf{On-demand observation improves cache reuse.}
The \texttt{screenshot} action lets the model omit screenshots when program output is sufficient, aiming to reduce unnecessary visual context. In the selected runs, cache hit rises from 75.6\% to 86.1\% for \qwenmax{} and from 61.9\% to 76.3\% for DeepSeek-V4.1-Flash, with higher task scores in both cases.

\textbf{Task-state tracking and action intent serve distinct goals.}
The \texttt{todo} interface supports task completion by maintaining a record of progress and unresolved work. Task score increases from 62.4 to 64.0 for \qwenmax{} and from 56.7 to 58.5 for DeepSeek-V4.1-Flash. In contrast, the \texttt{intent} annotation makes each program's purpose explicit so that real-device users can understand what the agent is doing. Its primary goal is user understanding rather than benchmark performance. Scores fall slightly.

\textbf{Cache reuse and context length matter for practical agent execution.}
As interaction histories grow, context policies must control their length while preserving cache reuse. The rolling screenshot buffer removes older images in batches, reducing how often prior context is modified. In the selected runs, cache hit improves by \textbf{16.0 and 13.4 percentage points} for \qwenmax{} and DeepSeek-V4.1-Flash, respectively. Micro-compaction instead replaces older tool outputs with short placeholders to limit context growth, at the potential cost of invalidating cached prefixes. Cache hit falls from 75.6\% to 68.6\% for \qwenmax{}, and from 61.9\% to 58.9\% for DeepSeek-V4.1-Flash. Neither context policy lowers task score in these selected comparisons.

\subsubsection{How Much Task Performance Can Money and Time Buy?}
\label{sec:money-time}

Final benchmark scores alone do not show how much performance is available under limited resources. We therefore examine monetary and time constraints using the same selected full-setting runs as the main evaluation on the offline subset of \osworldtwo{}.

\begin{figure}[t]
    \centering
    \includegraphics[width=0.95\linewidth]{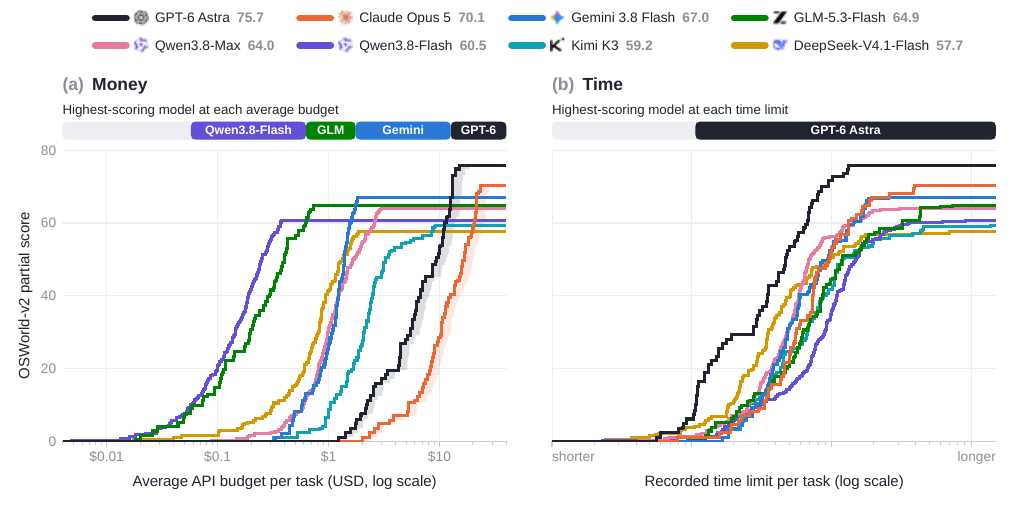}
    \caption{\textbf{Task performance under monetary and time constraints.}
    Mean partial scores on the 82-task offline subset of \osworldtwo{} under increasing per-task spending caps (a) and recorded time limits (b).
    Panel~(a) reports average expenditure, including unsuccessful attempts.
    Top bars identify the highest-scoring models; shading indicates cache-write cost bounds.
    Time comparisons are exploratory.}
    \label{fig:money-time}
\end{figure}

\textbf{Measuring Performance under Resource Limits.}
We retrospectively apply a common spending cap to every task. A trajectory contributes its recorded final partial score only if its full cost fits within the cap; otherwise, it contributes zero. 
We estimate spending by assuming that each attempt ends when its recorded trajectory finishes or its spending cap is reached, whichever comes first. Spending on unsuccessful attempts is also included.
Figure~\ref{fig:money-time}(a) plots the resulting score against average expenditure per task, not the cap itself. Panel~(b) applies the same scoring rule to a per-task time limit. Scores are averaged over all tasks. 

{Notably,} recorded wall-clock time depends on API bandwidth and rate limits, the number of concurrently active evaluation environments, provider-side queuing, and tool execution latency. These conditions were not controlled across models, limiting the comparability of the measurements. We therefore report the time curves as exploratory evidence and treat their trends as indicative rather than conclusive.

\textbf{Money and Time Favor Different Models.}
Under the assumed API prices, the highest-scoring model shifts from \qwenflash{} to GLM-5.3-Flash, Gemini 3.8 Flash, and eventually GPT-6 Astra as the monetary budget increases. For example, \qwenflash{} reaches 60.5 points at approximately \$0.38 per task, whereas GPT-6 Astra reaches the highest score of 75.7 at approximately \$15.26 per task. In the recorded-time comparison, however, GPT-6 Astra achieves higher scores at most time limits. 

\textbf{From Benchmark Scores to Resource-Aware Model Selection.}
In practice, the relevant question is not only which model achieves the highest benchmark score, but which delivers the most task performance within a given budget or an acceptable completion time. These curves make this distinction explicit: models with lower final scores may be preferable under tighter resource constraints.

\section{Real-Device Generalist CUA}
\label{sec:realmachine}
\vspace{-0.5em}
In Section~\ref{sec:model}, we study code execution and harness-to-model decision transfer on \osworldtwo{}, where every task runs in a controlled benchmark environment. Users, however, run agents on their own computers.
There, the environment is not initialized for the task, many application windows overlap, the user works at the same time, and requests change as the task goes on, so an agent that performs well on a benchmark may still struggle on a real device. 
This section first describes how real devices differ from benchmarks (Section~\ref{sec:gap}). It then presents the \sdkname{}, the programming interface through which the model operates different devices (Section~\ref{sec:sdk}), and the harness we built for real devices, organized into five layers with clear responsibilities (Section~\ref{sec:harness}).

\subsection{The Gap Between Benchmarks and Real Machines}
\label{sec:gap}

Controlled computer-use evaluations typically provide prepared task states, an initial instruction, and an environment reserved for the agent. Such settings simplify evaluation but do not fully capture the demands of operating a user's computer. We highlight five requirements that become especially important when the agent works alongside a user in an existing, changing environment.

\textbf{The user and agent share control.}
The user may continue working while the agent runs, so the agent cannot assume exclusive control of the pointer, keyboard, or focused window. It must account for user activity rather than attribute every change to its own actions. Where supported, it should prefer accessibility actions or background-window operations that leave the user's current interaction undisturbed. When these methods cannot complete an operation, foreground interaction requires coordination with the user before taking control.
 
\textbf{The screen keeps changing.}
Application windows open, close, and move, and one window can cover another. These changes may come from applications unrelated to the task and continue while the agent is reasoning. A coordinate or element reference that was valid when observed may therefore be wrong by the time the agent acts.

\textbf{User requests evolve during execution.}
Users may clarify an initial request, correct the agent's approach, or add constraints as work proceeds. The agent must keep track of the overall task goal and relevant progress across interruptions and turns, while revising its plan to reflect the user's latest instructions. 

\textbf{Observation and control require explicit targets.}
On a user's computer, the task-relevant application, window, or browser tab need not be in the foreground. A full-desktop screenshot shows the visible composition of windows, not the contents of every addressable target. The model must therefore choose both what to inspect and how to inspect it: a target-specific image, an accessibility tree, or a programmatic query may expose different aspects of the same target. For example, an agent inspecting a background product page may receive an unrelated login page if it requests a desktop screenshot rather than an image of the selected tab, as illustrated in Section~\ref{sec:realdevice-case-studies}.

Observation and control also impose different requirements. Being able to inspect a background target does not imply that every operation on it can run in the background. The model must select a supported control method, while the harness makes target identities, available capabilities, and foreground-input requirements explicit. Observations and actions should remain tied to their selected targets rather than silently following changes in desktop focus. These requirements link the model's observation and action choices to the resource bindings and capability declarations in the \sdkname{} (Section~\ref{sec:sdk}).

\textbf{Tasks begin in an existing user environment.}
Benchmark tasks typically start from a prepared state. A user's computer instead retains files, open tabs, application windows, and account sessions from prior use. Relevant materials may be mixed with unrelated content or similarly named files, requiring the agent to identify the correct inputs before acting. Expired sessions or missing permissions may also require user involvement before execution can continue.

These requirements mean that benchmark scores alone do not establish reliable operation on users' devices. We therefore study real-device capability alongside benchmark performance, rather than treating deployment as an implementation detail. The following sections describe our SDK and harness, and Section~\ref{sec:realdevice-task-results} reports preliminary real-device results.

\subsection{\sdkname{}}
\label{sec:sdk}
A computer-use task may span a mobile application, desktop tools, and a browser-based service. The \sdkname{} provides a common programming interface through which one agent can coordinate these environments. It supports both the reuse of common operations across platforms and the composition of cross-device workflows within a single task. Its programming model builds on code-execution-based computer-use interfaces \citep{openai2026cuaguide, openai2026codexcu}.

\begin{figure}[t]
    \centering
    \includegraphics[width=\linewidth]{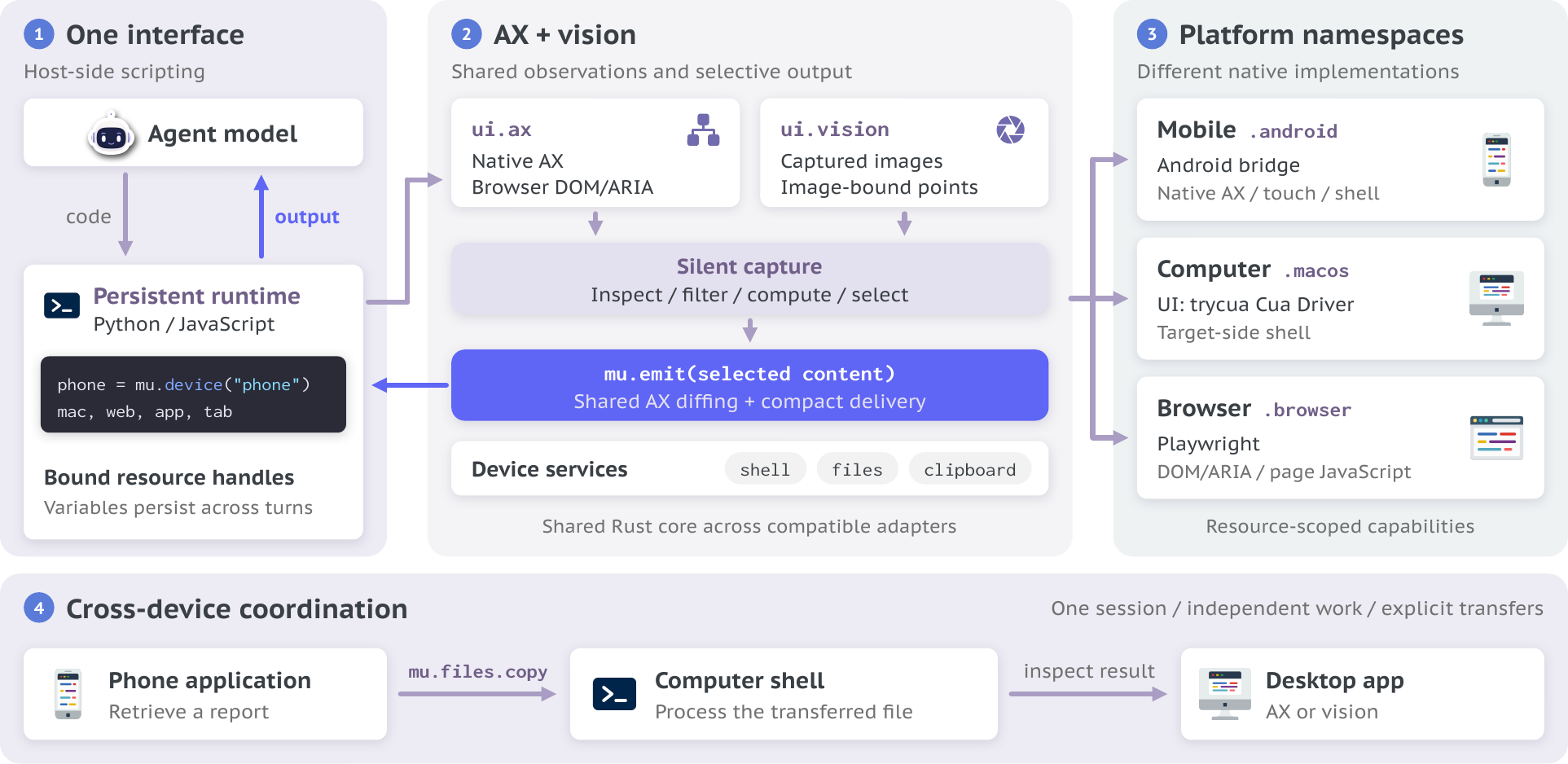}
    \caption{\textbf{Architecture of the \sdkname{}.} A common programming interface connects agent programs to computers, mobile devices, and browser environments, enabling cross-device workflows through platform-specific adapters.}
    \label{fig:sdk-architecture}
\end{figure}

\paragraph{Shared Operations with Explicit Targets.}
Common operations use a shared vocabulary, while platform adapters implement their native behavior. For example, the macOS adapter uses the unmodified Cua Driver from trycua \citep{trycua}. Each operation is scoped to an explicit resource, such as a device, application, window, or browser tab. Resource handles retain their targets independently, so selecting a new target does not redirect earlier bindings. This gives agent programs a consistent way to address multiple devices and applications without relying on a single implicitly selected target.

\paragraph{Cross-Device Coordination within One Task.}
Within one session, programs can combine observations and actions from multiple resources while reusing host-side variables and bindings across turns. The program executes in the host runtime, whereas \texttt{device.shell.run} executes commands on the selected device where supported. File transfers through \texttt{mu.files.copy} identify both source and destination resources. These mechanisms allow intermediate results from one environment to guide operations in another. As an illustrative workflow, an agent could retrieve a report through a phone application, transfer it to a computer for analysis, and enter the results into a browser-based service, coordinating all three stages within the same task.

\paragraph{Shared Observation Processing and Selective Delivery.}
The SDK exposes \texttt{ui.ax} for structured semantic observations and element-based actions, and \texttt{ui.vision} for images and coordinate-based interaction. Semantic state supports programmatic conditions and checks, while vision provides complementary information when semantic observations are insufficient. Adapters normalize native accessibility and browser semantic trees into common representations, allowing filtering and compact rendering to be reused across compatible platforms.

Observation capture is separate from delivery to the model. Calls such as \texttt{ui.ax.observe()} return data to the program, while \texttt{mu.emit} delivers selected observations or computed results to the model. A program can therefore inspect multiple targets, use structured results to determine subsequent operations, and return only the information needed for the next model decision. This separation supports cross-device execution without requiring a model call for every intermediate structured observation.

\paragraph{Explicit Platform Capabilities.}
A common interface does not imply identical capabilities across platforms. Each resource declares its supported operations and options. Common UI operations and device services share interfaces, while platform-specific features remain explicit through namespaces such as \texttt{.android}, \texttt{.macos}, and \texttt{.browser}. Operations such as Android navigation and browser page evaluation thus retain their own semantics, allowing programs to reuse common interaction patterns while selecting methods appropriate to each target.

\subsection{Real-Device Harness}
\label{sec:harness}
\begin{figure}[t]
    \centering
    \includegraphics[width=0.9\linewidth]{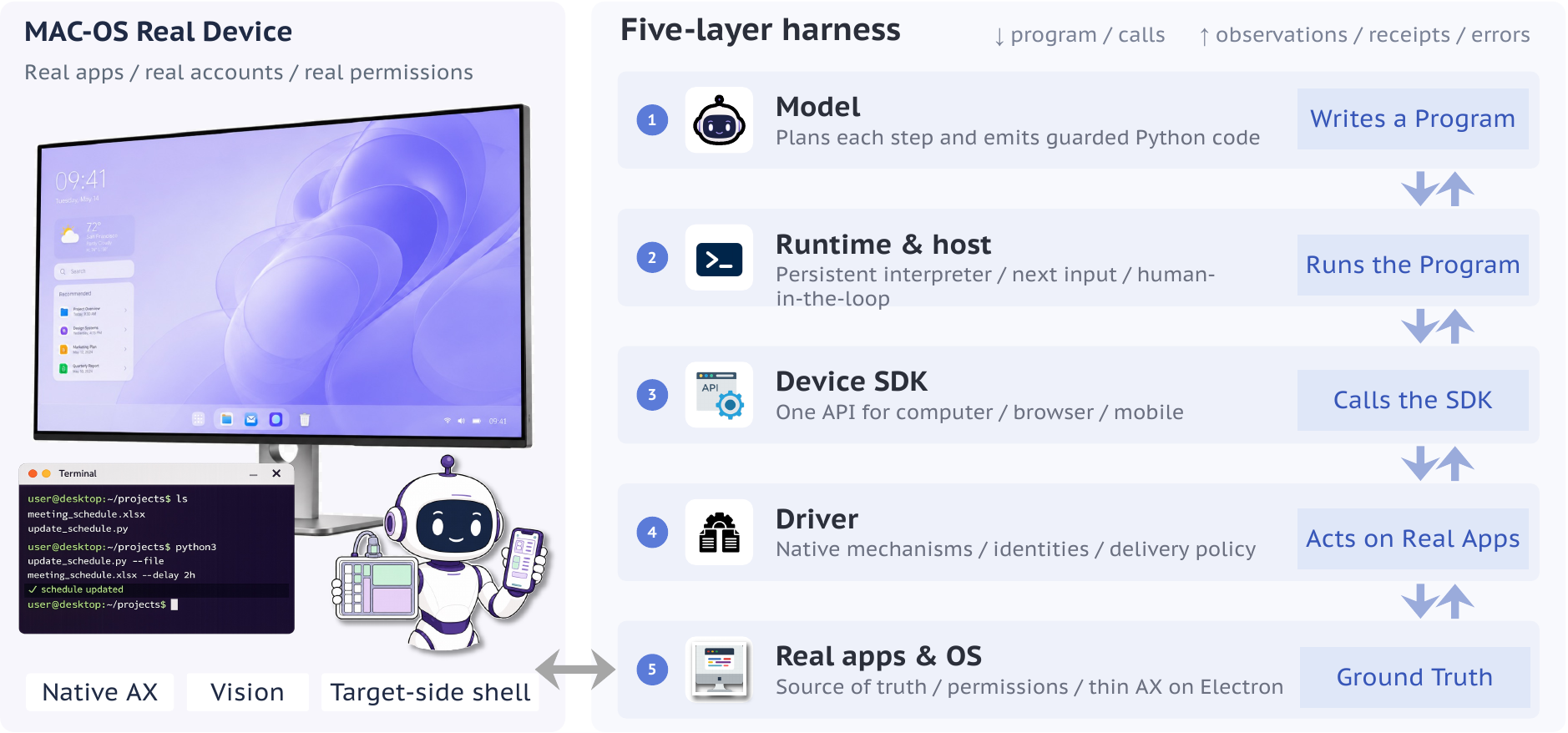}
    \caption{\textbf{Five-layer architecture of the real-device harness.} The runtime executes model-generated programs through the SDK and platform drivers to interact with applications and the operating system.}
    \label{fig:harness-layers}
\end{figure}

The \sdkname{} defines the programming interface for device observation and control. The real-device harness integrates this interface with the agent loop, program execution, and platform-specific mechanisms, enabling the model to work on a user's computer while retaining task state across turns.

\subsubsection{Five-Layer Architecture}
We separate the system into five layers (Figure~\ref{fig:harness-layers}). The model generates a program, the runtime executes it, and SDK calls are translated by the driver into operations on applications and the operating system. Explicit interfaces separate responsibilities and allow compatible components to be replaced independently. Detectable execution failures are reported where they occur; successful action delivery is not treated as evidence that the intended task-level effect has occurred.

\textbf{Model.}
Conditioned on the task, system prompt, and interaction history, the model decides how to observe, act, and organize its work. It selects targets and operations, maintains task state, and determines when to request user input. These decisions are expressed as Python programs against a common SDK, allowing device operations and programmatic checks to share the same control flow.

\textbf{Runtime.}
The runtime manages the agent loop and executes programs in a persistent interpreter. Variables, imports, and the working directory persist across turns, supporting incremental execution. It collects execution feedback and explicitly emitted observations, applies the configured context-retention policies, and constructs the next model input.

\textbf{Device SDK.}
The SDK binds observations and actions to explicit resources, such as devices, windows, and browser tabs (Section~\ref{sec:sdk}). Short element IDs written directly by the model must originate from observations delivered to it; programs may also obtain element references from fresh observations and use them within the same execution. Resource resets or rebinding invalidate associated references. These constraints establish the provenance and scope of action targets, but do not guarantee that the interface remains unchanged before execution.

\textbf{Driver.}
The driver translates SDK requests into supported native operations, including accessibility actions and coordinate-based input. It manages element identities and snapshot metadata, selects action-delivery mechanisms, and applies foreground or background delivery policies. Detectable native failures are returned as typed errors so that the agent can distinguish execution failures from outcomes requiring further inspection.

\textbf{Applications and Operating System.}
Applications and the operating system determine the available observations and control capabilities. Permissions, application state, and accessibility support constrain which operations can be performed. The harness must expose these limitations and use alternative interaction methods where supported.

\subsubsection{Responding to Real-Device Challenges}
The constraints discussed in Section~\ref{sec:gap} motivate four design choices that connect these layers.

\textbf{Minimizing Interference on Shared Machines.}
An agent operating alongside a user cannot assume exclusive control of the pointer, keyboard, or focused window. The harness therefore prefers element-level accessibility actions, followed by window-targeted coordinate input, and finally foreground input. Background delivery is used only where the target and platform support it; window-relative coordinates alone do not imply background execution. When a less intrusive method is unavailable or ineffective, the agent can escalate to a supported alternative, coordinating with the user before foreground input.

\textbf{Selecting Observations Explicitly.}
A full-desktop screenshot may show the user's activity rather than the agent's target, and visual input may be unnecessary during programmatic work. Following the model-initiated observation design in Section~\ref{sec:selfdirected}, the agent chooses when to observe and which target to inspect. A program can capture observations, process structured results, and emit relevant information within one execution; images are interpreted by the model when delivered in a subsequent input. Bounded history and image retention limit accumulated context. These mechanisms reduce unnecessary observations, but do not eliminate the need to refresh state when the interface changes.

\textbf{Managing Accessibility Information.}
Accessibility observations vary in size, stability, and coverage across applications. The harness filters irrelevant nodes and represents the retained hierarchy as a flat list linked by \texttt{parent\_id}. Element identities and snapshot versions support reference tracking, while \texttt{search}, \texttt{page}, and \texttt{read} retrieve relevant information from a retained snapshot without repeatedly capturing the window. When accessibility information is insufficient, visual observation provides a complementary source of evidence. The choice of action-delivery mechanism remains separate from the choice of observation modality.

\textbf{Maintaining Task Continuity through User Interaction.}
Real-device tasks may require clarification, authentication, or authorization that the agent cannot supply independently. The \texttt{ask\_user} interface supports these exchanges, including resolving ambiguous targets, requesting assistance with login or verification, and obtaining authorization for consequential actions. User responses and corrections become part of the ongoing task. The runtime preserves interpreter state, resource bindings, and interaction history, allowing execution to resume without rebuilding the workflow.

\begin{table}[!t]
\centering
\small
\setlength{\tabcolsep}{4pt}
\renewcommand{\arraystretch}{1.20}
\caption{\textbf{Real-device results on four macOS tasks.} Costs and completion use the same selected run per task (run 1). Costs are four-task means; Complete counts tasks judged complete. DeepSeek lacks a hotel trajectory, so its aggregates are unavailable.}
\label{tab:realmachine}
\begin{tabular*}{\linewidth}{@{\extracolsep{\fill}}lrrrr@{}}
\toprule[1.1pt]
\textbf{Model} & \multicolumn{3}{c}{\textbf{Interaction cost}} & \textbf{Task completion} \\
\cmidrule(lr){2-4}\cmidrule(lr){5-5}
 & Steps/task $\downarrow$ & Out K/task $\downarrow$ & Est. USD/task $\downarrow$ & Complete $\uparrow$ \\
\midrule
GPT-6 Astra & \cellcolor{realdeviceHighlight}\textbf{34.5} & \cellcolor{realdeviceHighlight}\textbf{12.9} & 18.290 & \cellcolor{realdeviceHighlight}\textbf{4/4} \\
Claude Opus 5 & 52.3 & 42.3 & 17.117 & 3/4 \\
Gemini 3.8 Flash & 36.0 & 31.9 & 0.662 & 2/4 \\
Qwen3.8-Max & 50.8 & 73.4 & 1.494 & 3/4 \\
Qwen3.8-Flash & 67.3 & 75.7 & \cellcolor{realdeviceHighlight}\textbf{0.120} & 2/4 \\
\bottomrule[1.1pt]
\end{tabular*}
\end{table}

\subsubsection{Preliminary Real-Device Results}
\label{sec:realdevice-task-results}

As our real-device benchmark is still under development, we study seven model configurations on four macOS tasks: workbook reasoning, PDF form filling, dietary-constrained shopping research, and hotel comparison, with up to three recorded runs per task and model. Table~\ref{tab:realmachine} summarizes five configurations, reporting mean interaction costs across all four tasks, including incomplete tasks. Steps count model responses, and output tokens include reasoning where recorded. Dollar costs are estimated from recorded response usage, cache hits, and public API prices.

\textbf{Execution Efficiency.}
Generation volume differs more sharply than interaction length. GPT-6 Astra averages 34.5 steps and 12.9K output tokens per task, compared with 50.8 steps and 73.4K tokens for Qwen3.8-Max: \textbf{1.5 times as many steps and 5.7 times as many output tokens}. Much of this difference lies in reported reasoning, which averages 2.6K tokens for GPT and 50.2K for Qwen. Claude Opus 5 and Gemini 3.8 Flash use 42.3K and 31.9K output tokens per task, respectively. Token usage and monetary cost need not rank models in the same order: under the archived pricing assumptions, Qwen3.8-Max costs \$1.49 per task, compared with \$18.29 for GPT-6 Astra. 

\paragraph{Task-Level Differences.}
On the workbook task, both selected runs satisfy the evaluation criteria, but GPT-6 Astra uses 22 steps and 8,783 output tokens, compared with 37 steps and 120,831 tokens for Qwen3.8-Max. Reported reasoning accounts for 1,487 and 92,974 tokens, respectively. On the PDF task, Qwen also uses significantly more output tokens (28,268 versus 10,013). The following case studies examine how observation choices and code execution contribute to the observed differences. With only four tasks and one run per model, these results characterize the evaluated configurations rather than establish a general model ranking.

\begin{figure}[tp]
    \centering
    \includegraphics[width=\linewidth]{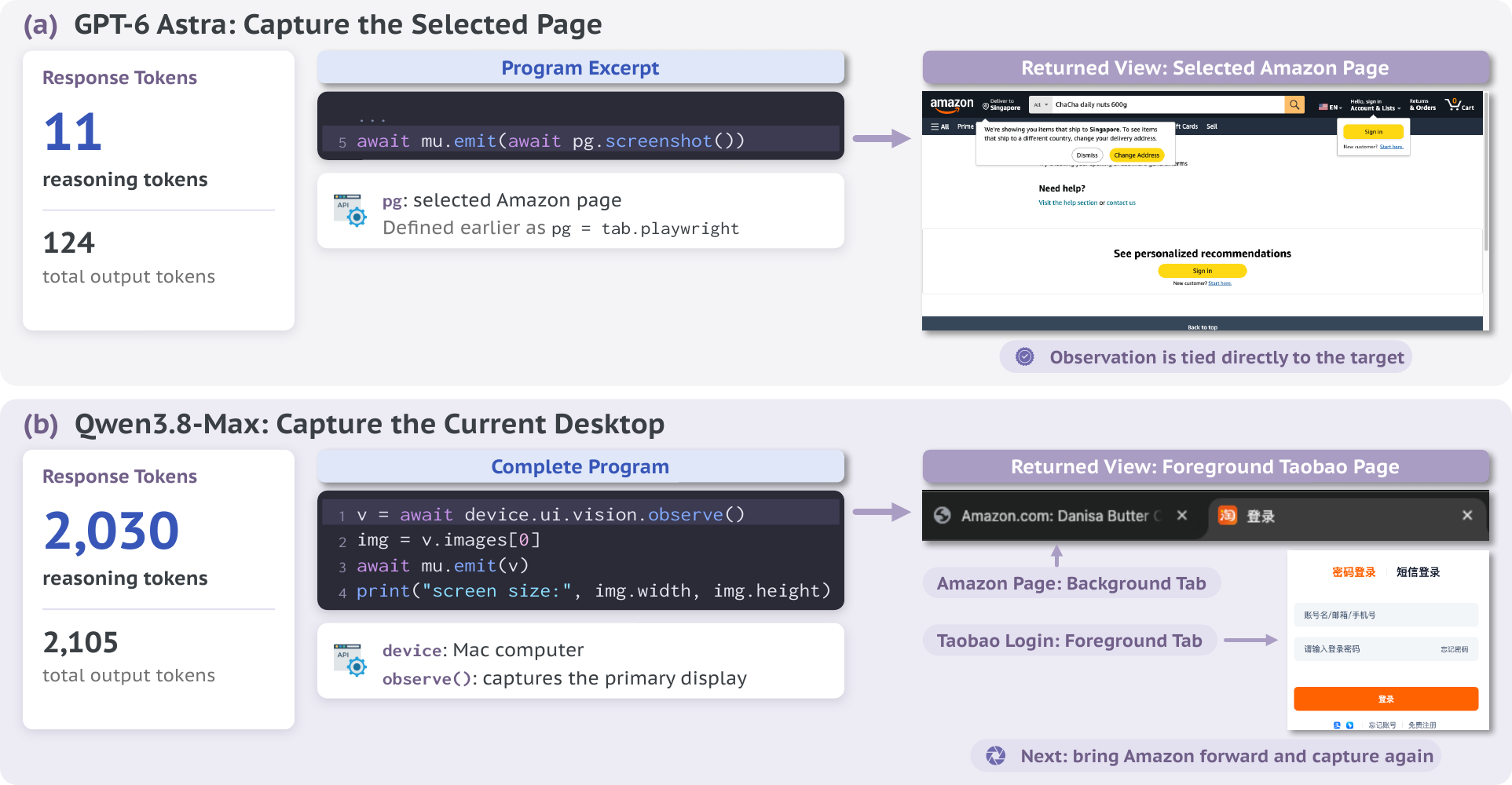}
    \caption{\textbf{Reasoning cost and observation-target selection.}
    GPT-6 Astra captures the selected Amazon page, while Qwen3.8-Max captures the foreground Taobao page and then switches to Amazon. Qwen3.8-Max uses substantially more reasoning tokens.}
    \label{fig:realdevice-observation}
\end{figure}
\subsubsection{Case Studies}
\label{sec:realdevice-case-studies}

\textbf{The gap in real-device execution extends from basic operation to completing an entire task.}
We compare GPT-6 Astra and Qwen3.8-Max through earlier low-effort trajectories under the same harness on real-device tasks, separate from the selected runs in Table~\ref{tab:realmachine}. The following cases show differences in the cost of routine operations, the selection of observation targets, and the ability to turn a task into an effective program.

\textbf{Large Differences in Reasoning Cost for Simple Operations.}
In a shopping task, Qwen produces a four-line program that captures and returns a desktop screenshot. The program returns an image, but the model response uses 2,105 output tokens, including 2,030 reasoning tokens. GPT also obtains a screenshot in another step of the same task, but its entire response uses only 124 output tokens, including 11 reasoning tokens (Figure~\ref{fig:realdevice-observation}). This contrast shows that a substantial reasoning-cost gap can arise even in a simple observation step. Such generation overhead can introduce substantial execution latency.

\textbf{Difficulty Adapting to Real-Device Conditions.}
On a real device, the page an agent needs to inspect may not be in the foreground. In the shopping task, Qwen is working with an Amazon browser object but requests a desktop screenshot, which shows the foreground Taobao login page. It then brings Amazon forward and captures again. GPT instead requests a screenshot directly from its selected browser page. Both models use screenshots; the contrast concerns observation scope, illustrating why the intended task target cannot be assumed to coincide with the foreground interface.

\begin{figure}[tp]
    \centering
    \includegraphics[width=\linewidth]{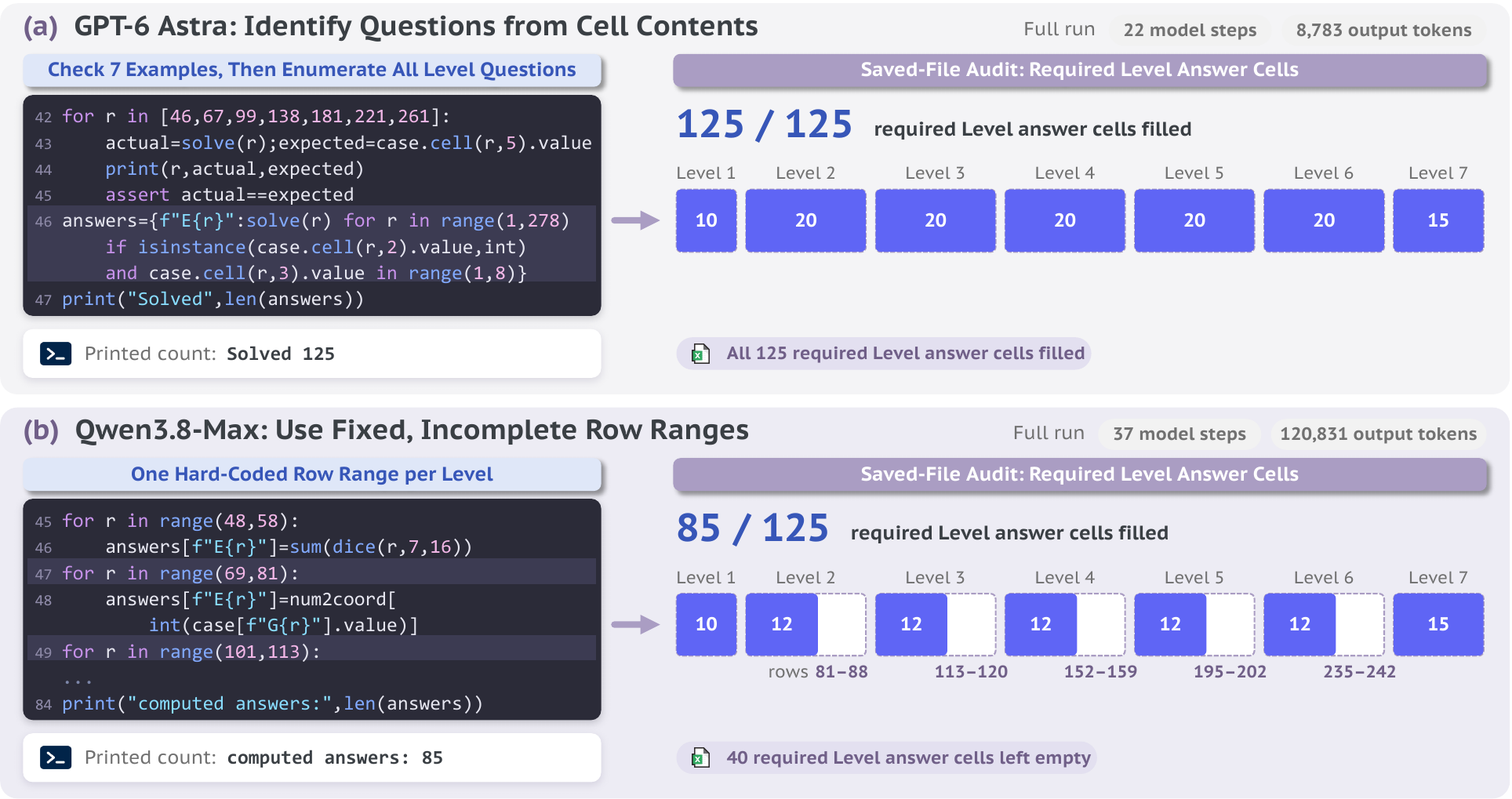}
    \caption{\textbf{Task coverage in code execution.}
    GPT-6 Astra fills all 125 required Level answer cells, while Qwen3.8-Max leaves 40 empty. }
    \label{fig:realdevice-programs}
\end{figure}

\textbf{A Gap in Task-Level Code Execution.}
A workbook task reveals a difference in how models translate task requirements into programs. GPT-6 Astra builds a reusable solver, checks it against seven supplied examples, and identifies question rows from the workbook contents. Its program generates answers for all 125 questions. \qwenmax{} also uses functions and loops, but processes fixed row ranges that cover only 85 questions. Although its program reports this count, \qwenmax{} does not check for omitted questions. An audit of the saved workbooks confirms that GPT fills all 125 required Level answer cells, while \qwenmax{} leaves 40 empty (Figure~\ref{fig:realdevice-programs}). The gap is therefore not simply whether a model can write executable code with loops and conditions, but whether it can use code correctly to cover the full task and check for incomplete work.

\section{Roadmap}
\label{sec:roadmap}
\vspace{-0.5em}
\projname{} aims to make generalist computer use a shared research effort in which models, execution systems, and real-device evaluation advance together. This first release provides an initial basis for that work.

\subsection{From Design Validation to Model Release}
\label{sec:roadmap-release}
The model-side contribution of this report is \textbf{empirical validation of agent design using existing models}, rather than a newly trained model. We study code execution and harness-to-model decision transfer, and release implementations, prompts, and evaluation trajectories alongside preliminary versions of the real-device harness and cross-device SDK.

\textbf{A GCUA model is the central addition planned for the next release.}
We aim to unify computer and browser use and turn the designs studied here into learned capabilities: coordinating visual and programmatic operations through code, requesting observations when needed, maintaining task state throughout execution, and particularly enhancing the capability in real-device scenario.

\subsection{Research Priorities}
\label{sec:roadmap-next}
\label{sec:roadmap-research}
\textbf{More Efficient Reasoning and Execution.}
We will optimize the model's thinking, making reasoning more concise for routine operations and mitigating overthinking. On the harness and infrastructure side, we will improve code execution, context management, and infrastructure optimization. We will also explore new system architectures and other system-level solutions for efficiency.

\textbf{Reliable Work on Users' Devices.}
We plan to extend the harness and SDK toward tasks spanning computers, browsers, and potentially mobile devices. Priorities include reliable target selection, recovery from execution failures, and task continuity when users interrupt or change their requests. We will also investigate which additional workflow decisions should move from fixed harness rules to the model.

\textbf{Real-Device Environments and Benchmarks.}
We will develop evaluation environments and benchmarks on real devices, particularly macOS, covering a wider range of applications, user interactions, and changing device states. The goal is to measure whether agents complete useful work under real operating conditions, together with the time and resources required.

\subsection{How to Contribute}
\label{sec:roadmap-contribute}
We welcome contributions to both the implementation and the research agenda:
\begin{itemize}[leftmargin=*]
    \item \textbf{Improve the Harness and SDK.} Contribute platform adapters, more reliable observation and control, execution optimizations, or real-device testing and documentation.
    \item \textbf{Build Real-Device Evaluation.} Contribute environments, realistic tasks, evaluation criteria, and benchmark infrastructure, including execution traces that expose current limitations.
    \item \textbf{Propose Research Goals.} Identify workflows agents should support, capabilities worth developing, or questions to investigate through the released systems and trajectories.
\end{itemize}
These priorities guide planned releases without fixing a delivery timeline. The agenda will evolve with experimental evidence and community proposals, including directions beyond those listed here. Our goal is for the community to help shape not only how GCUAs are built, but also what they should be able to do and how progress should be measured.

\section{Related Work}
\label{sec:related}
\vspace{-0.5em}

\textbf{Frontier Computer Use.} OpenAI positions GPT-6 Astra for difficult end-to-end work and describes computer use as a new frontier in speed, accuracy, and safety. It recommends code execution for computer-use tasks while retaining the structured \texttt{computer} tool as an alternative \citep{openai2026gpt6astra, openai2026cuaguide}. In Codex, Computer Use can see and operate graphical user interfaces on macOS and Windows, support workflows spanning multiple applications, and, on macOS, run scoped background tasks while the user continues working elsewhere \citep{openai2026codexcu}. Anthropic also provides computer use in Claude Cowork and Claude Code: Claude first reaches for the most precise tool, such as connectors to services like Slack or Google Calendar, and can directly control the browser, mouse, keyboard, and screen when no connector is available. Its Dispatch feature further lets users assign tasks from a phone and open the completed work on a computer \citep{anthropic2026dispatch}. Despite these capabilities, the public releases leave key implementation details and execution trajectories unavailable, making their system designs difficult for the community to reproduce and study systematically.

\textbf{Computer-Use Agents.} Prior open CUA models remain largely GUI-centric. UI-TARS and UI-TARS-2 \citep{uitars, uitars2}, OpenCUA \citep{opencua}, ScaleCUA \citep{scale_cua}, EvoCUA \citep{evocua15}, and Fara \citep{fara} release open models for computer use and report strong results on relevant benchmarks. Recent work has begun to explore hybrid GUI--programmatic execution: CoAct-1, ComputerRL, UltraCUA, ToolCUA, and StateAct combine GUI control with code, APIs, tool calls, or program-state operations \citep{coact1, computerrl, ultracua, toolcua, stateact}; RecreationWorld and WeaveBench respectively provide environments and benchmarks for such interleaved trajectories \citep{recreationworld, weavebench}. Our previous work, \qwenui{}, introduced a unified GUI--CLI action space with batched actions, a real-device mobile runtime, and a harness for proactive cross-platform execution \citep{qwenuiagent, maiui}. To capture this broader setting, following the system-level perspective of Agents' Last Exam \citep{agentslastexam}, we use the term \textbf{Generalist Computer-Use Agent} (GCUA). We define a GCUA as \emph{an agent system that completes diverse computer tasks through visual interaction and programmatic access, choosing how to observe, act, and organize its work as each task requires}.

\textbf{Open Computer-Use Infrastructure.} Several projects provide open harnesses for computer-use agents. The \texttt{cua} project provides cross-platform desktop drivers and sandbox SDKs \citep{trycua}. The Agent S series provides open orchestration frameworks with memory, visual grounding, and increasingly code-based execution \citep{agents, agents2, agents3}. Open Computer Use provides an MCP-wrapped desktop-control runtime, including macOS control through Accessibility \citep{opencodexcu}. These are complementary to our work. \projname{} is an ongoing project to open a full GCUA research stack that couples a trained model with a real-device harness, a cross-device SDK, and execution trajectories.

\section{Conclusion}
\label{sec:conclusion}
\vspace{-0.5em}
We presented the first release of \projname{}, an ongoing project to develop and open the full stack for generalist computer use. Our GCUA formulation centers on three model decisions: what to observe, how to act, and how to organize work. Code execution lets models use intermediate results to coordinate GUI and programmatic operations and apply their existing programming capabilities to computer interaction. Harness-to-model decision transfer complements this design by moving selected observation and task-state decisions from fixed harness rules to the model. On the offline subset of \osworldtwo{}, the full OpenGCUA configuration improves task scores across all eight evaluated models, raising mean partial score by 19.6 points and reducing reported steps per task by an average of 53.0\% relative to batched GUI-only execution. Ablations show additional benefits from code execution with the same GUI and CLI access, indicating that execution design matters beyond interface access.

We release our implementation, prompts, and evaluation trajectories, alongside preliminary versions of a real-device harness and cross-device SDK. Evaluations on four macOS tasks reveal substantial differences in reported reasoning-token usage and task completion under the same harness. This report studies GCUA designs rather than introducing a newly trained model. OpenGCUA will continue through further research and releases on GCUA models, reasoning and execution efficiency, and real-device environments and benchmarks.

\clearpage
\bibliography{biblio}
\bibliographystyle{colm2024_conference}

\clearpage
\appendix
\addtocontents{toc}{\protect\setcounter{tocdepth}{1}}
\section{Evaluation Details}
\label{app:eval}
\lstdefinestyle{appendixcode}{
    basicstyle=\ttfamily\scriptsize,
    columns=fullflexible,
    keepspaces=true,
    breaklines=true,
    breakatwhitespace=false,
    showstringspaces=false,
    numbers=none,
    frame=single,
    rulecolor=\color{black!25},
    framerule=0.3pt,
    backgroundcolor=\color{black!2},
    aboveskip=5pt,
    belowskip=6pt,
    xleftmargin=4pt,
    xrightmargin=4pt
}
\textbf{Metrics and Aggregation.}
We evaluate the 82-task offline subset of \osworldtwo{} v2.1. Task score is the mean evaluator partial score, expressed on a 0--100 scale. Full-task success is the percentage of tasks with a normalized evaluator score greater than 0.999999. Steps and token counts are averaged over all selected tasks, including failures. Within each run, we retain the latest attempt with an evaluator result for each task, not the highest-scoring attempt. Cross-model summaries give each model equal weight: score gains are averaged in points, and relative resource reductions are calculated within each model before averaging.

The offline subset is defined by \nolinkurl{evaluation_examples/test_v2_offline_no_internet.json}. It excludes the 26 tasks requiring external Internet access, while retaining local web applications and desktop tasks.

\section{Real-Device Evaluation Details}
\label{app:realdevice-evaluation}

\subsection{Task Descriptions}
\label{app:realdevice-task-descriptions}

\paragraph{Workbook Reasoning.}
The input workbook contains race rules, a track layout, seven worked examples, and 125 Level questions. The agent must compute all Level answers and the requested bonus answers, enter them in the designated cells, and save a completed copy without altering the original.

\paragraph{PDF Form Filling.}
The inputs comprise a blank four-page Schengen application form and synthetic identity, employment, travel, and accommodation records. The agent must fill all applicable fields supported by these materials and save a new PDF with all entered information visible and correct.

\paragraph{Shopping Research.}
Given 22 snack candidates and a dietary-restriction list, the agent must verify exact products on Amazon, falling back to official brand or original product pages when necessary. The DOCX must cover every candidate, with evidence for purchase or replacement recommendations and explicit uncertainty about unverified ingredients or allergens. No purchase is permitted.

\paragraph{Hotel Comparison.}
The agent must compare Pullman Paris Montparnasse and InterContinental Paris Le Grand for two adults over December 20--30, 2026, with one hotel change and at least three nights at each hotel. It lists ten feasible date plans and researches four prescribed plans through the official sites, Expedia, and Booking.com. A four-sheet workbook must document tax-inclusive EUR prices, refundability, breakfast for two on December 23--28, and source limitations. No reservation is permitted.

\FloatBarrier
\Needspace{10\baselineskip}
\subsection{Task Instructions and Materials}

\label{app:realdevice-materials}

The following are full English translations of the original task instructions. The snack-document title and corresponding filename are translated into English. The original prompts and input files are included in the accompanying task-material bundle.

\subsubsection{Workbook reasoning and editing}

\label{app:realdevice-instruction-t1}

\paragraph{Input Materials.} Elimination\_\allowbreak{}Race\_\allowbreak{}to\_\allowbreak{}Las\_\allowbreak{}Vegas.xlsx, containing the Case and RaceTrack sheets. The source contains 125 Level 1-7 questions, seven worked examples, and five bonus questions.

\paragraph{Task Instruction.}

\begin{quote}\small\setlength{\parskip}{3pt}\setlength{\parindent}{0pt}\setlength{\emergencystretch}{2em}

Please use the file "Elimination\_Race\_to\_Las\_Vegas.xlsx" located on your desktop to complete the "Jade's Community Game! (Everyone Welcome)" challenge at https://play.excel-esports.com. Follow the rules provided in the workbook and on the challenge page to solve all the questions, enter your answers into the designated fields on the website, and submit them. Finally, display the results page and report both your actual score and the maximum possible score.

\end{quote}

\begin{figure}[!htb]
\centering
\includegraphics[width=\linewidth,height=0.32\textheight,keepaspectratio]{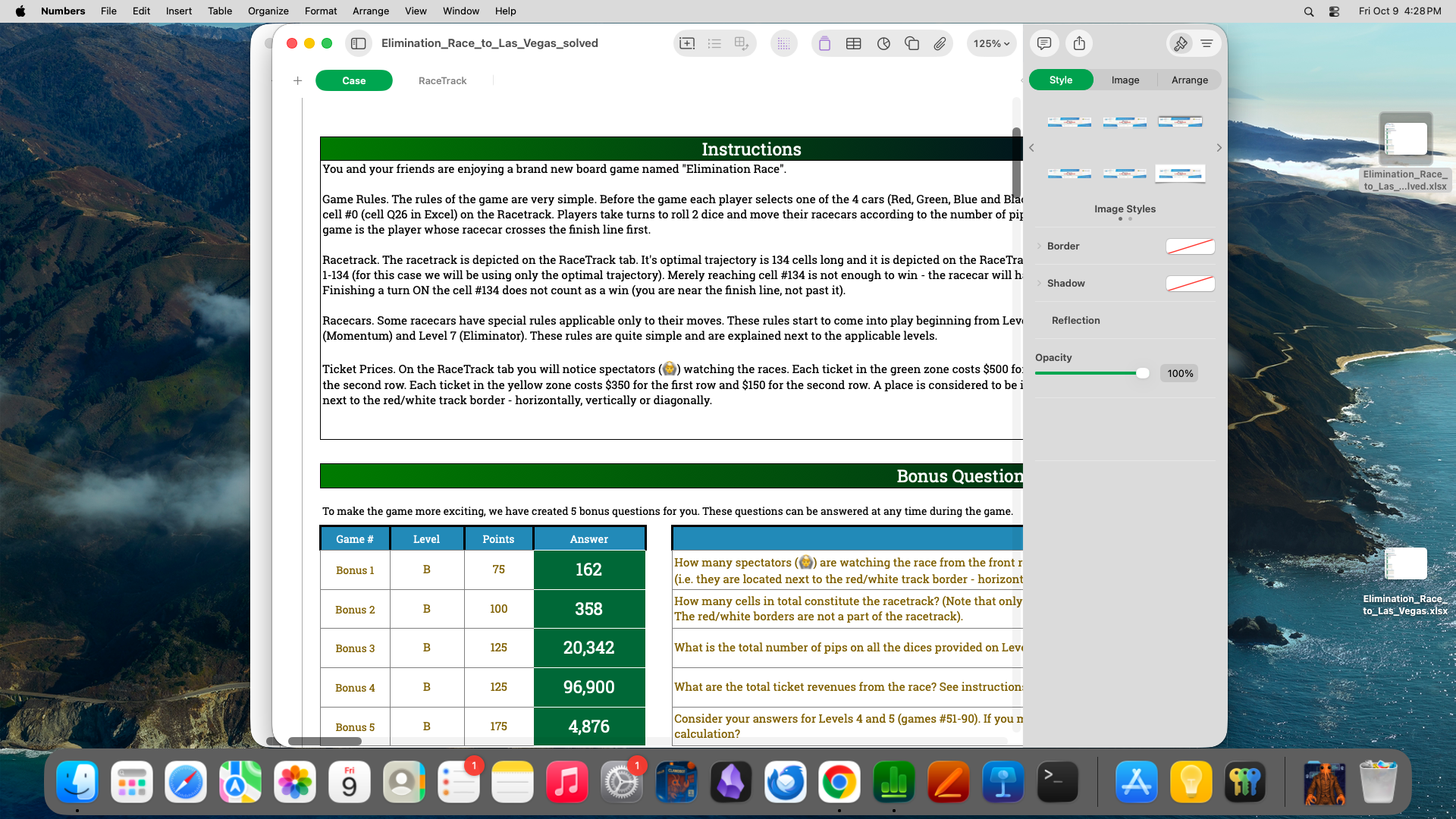}
\caption{Workbook interface during the Claude Opus 5 run. The view shows the rules and answer cells. Displayed answers are model outputs, not part of the initial workbook.}
\label{fig:realdevice-material-workbook}
\end{figure}

\FloatBarrier
\Needspace{10\baselineskip}

\subsubsection{Document-grounded PDF form filling}

\label{app:realdevice-instruction-t3}

\paragraph{Input Materials.} visa task.zip: a blank four-page Schengen form, passport and ID information, an application worksheet, an employment certificate, a flight itinerary, and hotel bookings. The identity workbook explicitly labels its records as synthetic test data.

\paragraph{Task Instruction.}

\begin{quote}\small\setlength{\parskip}{3pt}\setlength{\parindent}{0pt}\setlength{\emergencystretch}{2em}

Please use the application materials found in the \texttt{visa task.zip} file on the desktop to complete the blank Schengen visa application form, \texttt{CS\_14076-05\_EN\_05.pdf}. Review every field in the form and populate it with all clearly identifiable information provided within the \texttt{visa task.zip} package—including the applicant’s personal details, passport information, travel itinerary, accommodation, transportation, occupation, and any other relevant data. Once finished, save the completed form as \texttt{CS\_14076-05\_EN\_05\_filled.pdf}. The final PDF file must be accessible and retain all the information you have entered.

\end{quote}

\begin{figure}[!htb]
\centering
\includegraphics[width=\linewidth,height=0.48\textheight,keepaspectratio]{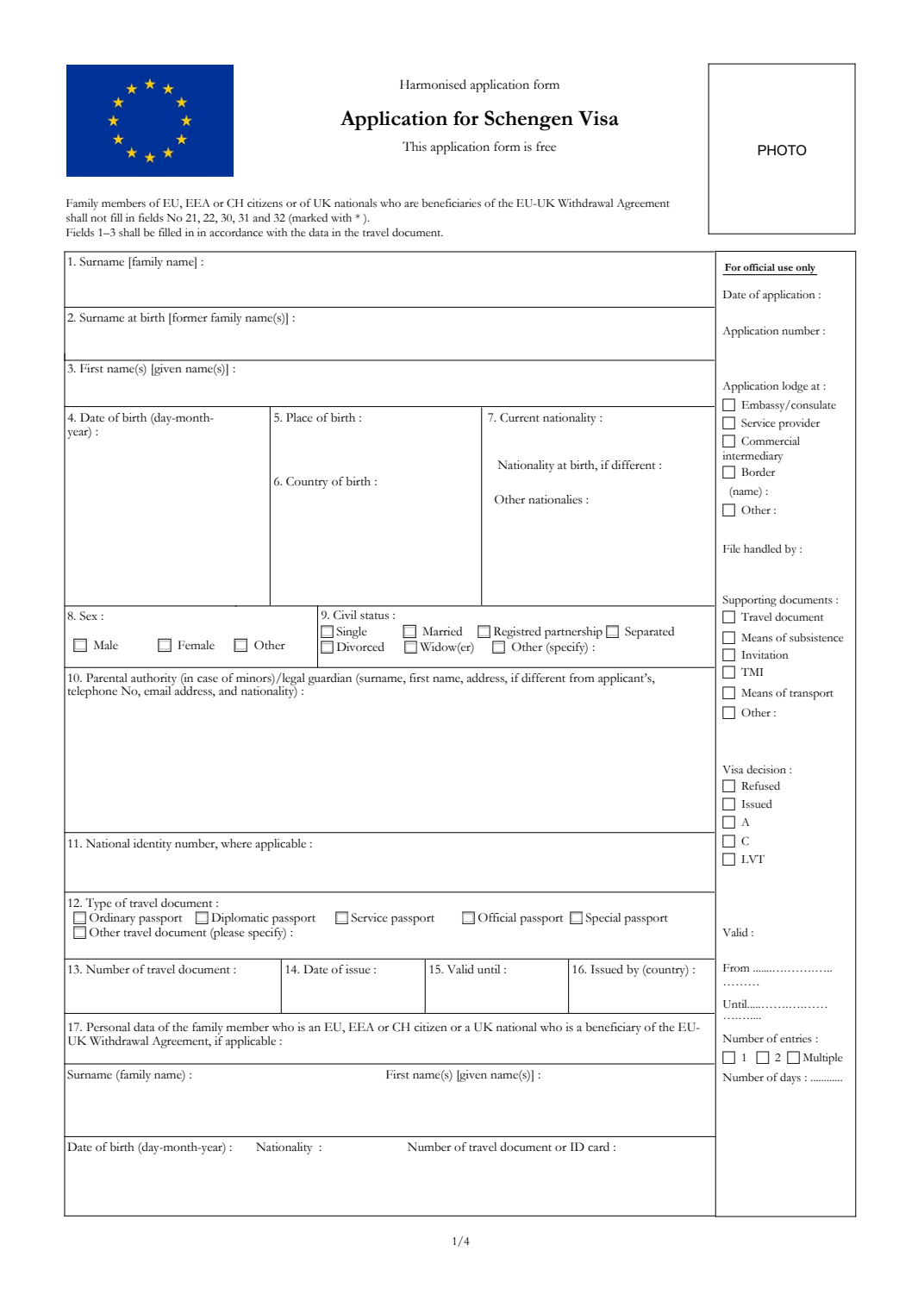}
\caption{First page of the supplied blank application form. Completed forms and applicant details are not shown.}
\label{fig:realdevice-material-pdf}
\end{figure}

\FloatBarrier
\Needspace{10\baselineskip}

\subsubsection{Evidence-grounded online shopping research}

\label{app:realdevice-instruction-t4}

\paragraph{Input Materials.} snake filter task.zip: Combined\_\allowbreak{}Snacks (22 candidates, with original product names and links) and Dad\_\allowbreak{}No\_\allowbreak{}Go\_\allowbreak{}List (dietary restrictions and allergen exclusions). Live sources are Amazon, official brand pages, or the original candidate product pages.

\paragraph{Task Instruction.}

\begin{quote}\small\setlength{\parskip}{3pt}\setlength{\parindent}{0pt}\sloppy

Please review the 22 snacks on the candidate list by cross-referencing them with the \texttt{Combined\_Snacks} and \texttt{Dad\_No\_Go\_List} files found in the \texttt{snake filter task.zip} folder on your desktop.

Since the candidate list does not provide full ingredient lists, please look up each item on Amazon. Use the ingredient list, allergen information, and product description provided on the Amazon product page—in conjunction with the restrictions outlined in the \texttt{Dad\_No\_Go\_List}—to determine whether each item is suitable for purchase.

Create a Word document titled ``Snacks Dad and I Can Share'' and categorize all 22 candidate items into the following three groups:
\begin{enumerate}
\item Safe to purchase as is
\item Requires a change in flavor, size, or other adjustments before it can be purchased
\item Not recommended for purchase
\end{enumerate}

For each item, you must include:
\begin{itemize}
\item Number
\item Product Name
\item Assessment Result
\item Rationale for the assessment
\item If adjustments are required, specify the necessary changes (e.g., flavor, size, or other modifications)
\end{itemize}

If the ingredients or allergen information for a product cannot be verified via its Amazon product page, do not rely on the product name or general assumptions; instead, clearly label it as ``Ingredient information unverified.''

Ensure that all 22 candidate items are included and that each item is assigned to exactly one category.

Include the following disclaimer at the end of the document:

Prices, formulations, and stock availability are subject to change. Always check the actual product packaging labels and warnings regarding cross-contact before placing an order or consuming the product.

\end{quote}

\begin{figure}[!htb]
\centering
\includegraphics[width=\linewidth,height=0.70\textheight,keepaspectratio]{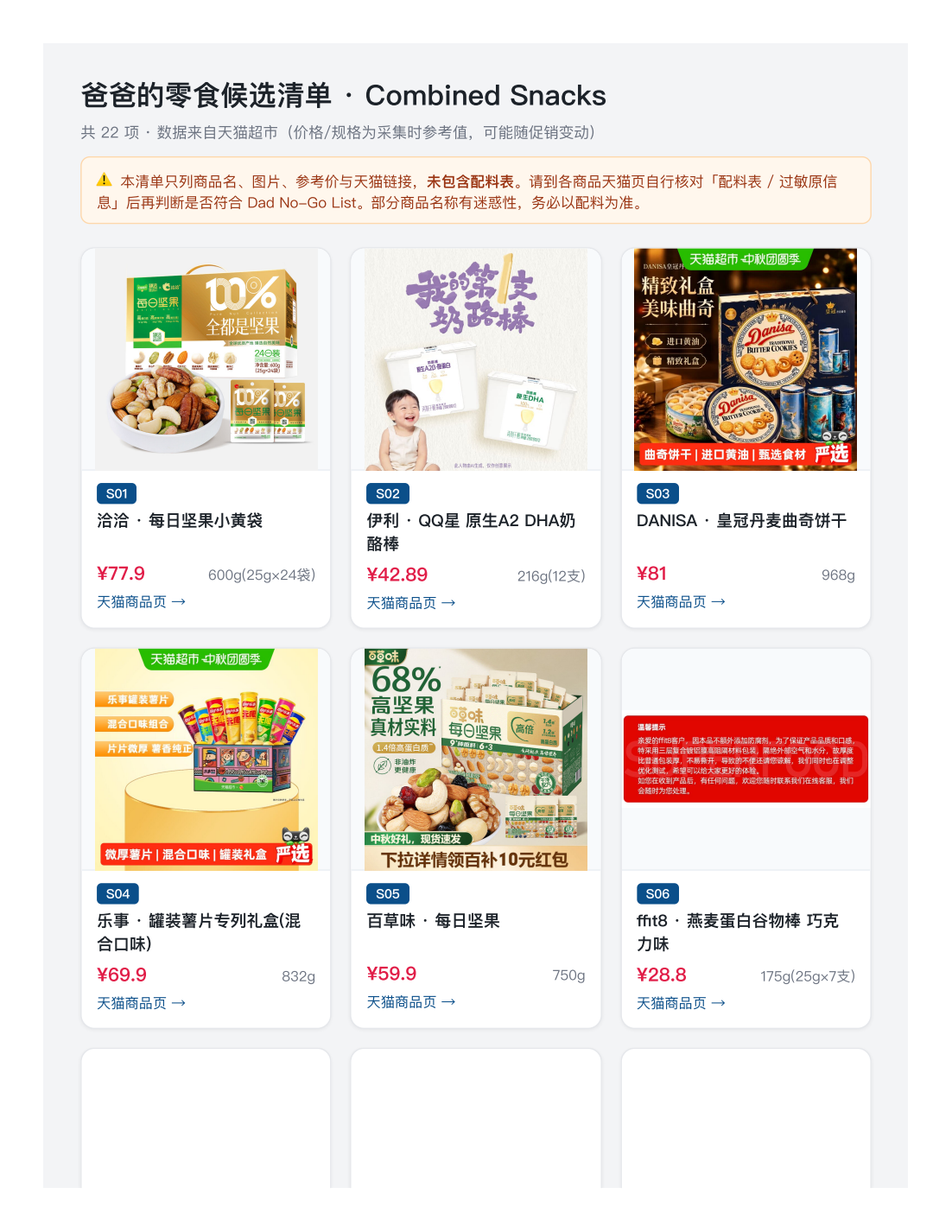}
\caption{First page of the supplied Combined\_\allowbreak{}Snacks candidate list, showing product images, package descriptions, and original listing links. The screenshot retains the source language.}
\label{fig:realdevice-material-shopping}
\end{figure}

\FloatBarrier
\Needspace{10\baselineskip}

\subsubsection{Constrained live hotel comparison}

\label{app:realdevice-instruction-t5}

\paragraph{Input Materials.} A text-only request for a Paris itinerary and a four-sheet comparison workbook. Prices and policies must come from the live Accor, IHG, Expedia, and Booking.com websites, not bundled rate cards.

\paragraph{Task Instruction.}

\begin{quote}\small\setlength{\parskip}{3pt}\setlength{\parindent}{0pt}\setlength{\emergencystretch}{2em}

I'm planning a Paris trip from Dec 20 to Dec 30, 2026 (check-in Dec 20, check-out Dec 30, 10 nights total) for 2 adults. I want to split the stay between exactly two hotels:

\begin{itemize}
\item \textbf{Pullman Paris Montparnasse}\\
Official site: \url{https://all.accor.com/a/en/offers/meapac/pullman.html?merchantid=ppc-pul-goo-apac-en-pmax&sourceid=aw-cen&utm_source=google&utm_medium=cpc&utm_campaign=ppc-pul-goo-apac-en-apac-pmax-aw-cen&utm_term=goo&utm_content=en-apac-all-all&wiz_campaign=24047123880&gad_source=1&gad_campaignid=24047157582&gbraid=0AAAABC20MbFGIzWYbNya4iMdEd7J85MfU&gclid=CjwKCAjww-3VBhAcEiwAwUUIu9eC4_lOMg5Q6UoljvjRNlUdcS3bBq5DHLxRAKHOZoVW10488aIrYBoCu3sQAvD_BwE}

\item \textbf{InterContinental Paris - Le Grand}\\
Official site: \url{https://www.ihg.com/intercontinental/hotels/us/en/paris/parhb/hoteldetail}
\end{itemize}

Also compare rates on \textbf{Expedia} and \textbf{Booking.com}.

Constraints:

\begin{enumerate}
\item Stay at each hotel for at least 3 consecutive nights, with only ONE hotel switch (Hotel A for \(N\) nights then Hotel B for \(10-N\) nights, N from 3 to 7, in either order).
\item 1 room, 2 adults, queen bed. If queen is unavailable, use king and clearly flag it.
\item Fully refundable / free-cancellation rates only.
\item Breakfast for 2 must be included on the mornings of Dec 23 through Dec 28 inclusive. Therefore, the nights of Dec 22 through Dec 27 must be booked on a breakfast-included rate. Other nights may use room-only refundable rates if cheaper.
\end{enumerate}

Task:

\begin{itemize}
\item Enumerate every valid hotel-order × switch-date × booking-site combination that satisfies the constraints.
\item Check the two official hotel sites above, Expedia, and Booking.com.
\item For each combination, record nightly rates, taxes/fees, cancellation deadline, breakfast coverage, and any verified direct-booking perks such as member rates, loyalty points, upgrades, hotel credits, or working promo codes.
\item Do not assume an unavailable rate, perk, promo code, breakfast inclusion, or cancellation policy. Record only information that can be verified from the booking page.
\end{itemize}

Output: create an Excel file (.xlsx) with:

\begin{itemize}
\item Sheet 1 \textbf{``All combinations''}: one row per valid combination. Columns = Rank | Hotel 1 | Hotel 1 dates | Hotel 1 nights | Hotel 1 site | Hotel 1 total EUR | Hotel 2 | Hotel 2 dates | Hotel 2 nights | Hotel 2 site | Hotel 2 total EUR | Grand total EUR | Grand total USD | Room type | Refundable until | Breakfast nights covered | Perks / promo codes | Notes. Sort by Grand total EUR ascending. Highlight the cheapest row in green and bold; highlight the cheapest single-site option (both hotels booked on the same site) in yellow.
\item Sheet 2 \textbf{``Rate detail''}: nightly price per hotel × date × site × rate type (room-only vs breakfast, refundable), including taxes/fees where available, so the combination totals can be verified.
\item Sheet 3 \textbf{``Summary''}: the lowest-cost valid arrangement in plain text — hotel order, exact dates, which site to book each hotel on, total price, and cancellation deadlines — plus the next 3 lowest-cost valid alternatives.
\end{itemize}

Do NOT book anything. Save and return the completed \texttt{.xlsx} file.

\end{quote}

\begin{figure}[!htb]
\centering
\includegraphics[width=\linewidth,height=0.27\textheight,keepaspectratio]{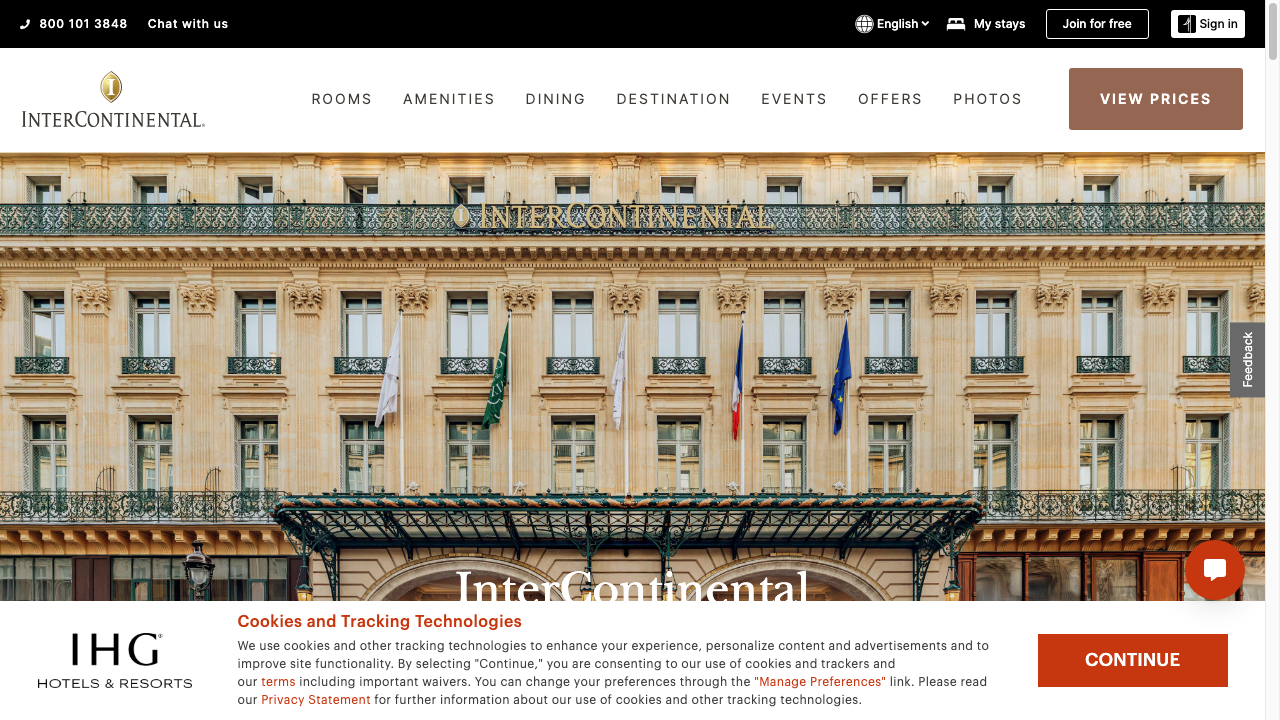}
\caption{InterContinental Paris Le Grand website observed during the GPT-6 Astra run.}
\label{fig:realdevice-screen-hotel}
\end{figure}

\Needspace{12\baselineskip}
\section{Prompts}
\label{app:prompts}
\lstdefinestyle{appendixprompt}{
    basicstyle=\ttfamily\fontsize{8}{9.5}\selectfont,
    language={},
    columns=fullflexible,
    keepspaces=true,
    breaklines=true,
    breakatwhitespace=false,
    breakautoindent=false,
    breakindent=1em,
    showstringspaces=false,
    showspaces=false,
    showtabs=false,
    numbers=none,
    frame=single,
    rulecolor=\color{black!25},
    framerule=0.3pt,
    framesep=5pt,
    backgroundcolor=\color{black!2},
    linewidth=\linewidth,
    xleftmargin=5.3pt,
    xrightmargin=5.3pt,
    aboveskip=4pt,
    belowskip=8pt
}
\newenvironment{ogcuapromptexcerpt}[1]{%
    \par\addvspace{5pt}\noindent\begin{minipage}{\linewidth}%
    \textbf{#1}\par
}{\end{minipage}\par\addvspace{3pt}}

We first present verbatim excerpts for code execution and harness-to-model decision transfer, then provide the complete OpenGCUA system prompt for the \texttt{all} setting. Only the environment login credential is redacted; the instructions and examples are preserved as recorded.

\subsection{Code Execution}
\label{app:prompt-code}
We reproduce the code-execution baseline's execution contract, PyAutoGUI interface, three code examples, and response format, retaining the original prompt headings. The PyAutoGUI definition is also used in the full prompt (Section~\ref{app:prompt-full}). The examples illustrate GUI interaction, conditional shell execution, and loops over GUI operations.

\begin{lstlisting}[style=appendixprompt]
## Code execution
The script in your `<code>` block runs top-to-bottom as a NEW Python process each step. Nothing carries over from a previous step - variables, imports, functions, working-directory changes, shell state, subprocess handles and separate `bash()` calls are all independent; only external state such as files, application state and surviving background system processes persists. Everything the script prints (stdout, stderr, `bash()` echo, tracebacks) is returned in the next observation, truncated to 3200 characters (first 1200 + last 2000). The script is killed after 300 seconds; an uncaught exception stops it and its traceback is returned. `</code>` is a reserved delimiter: if a script must emit that exact string, build it indirectly.

These names are predefined (no import needed): `pyautogui` (coordinate-aware, see below), `time`, `os`, `sys`, `re`, `json`, `subprocess`, `Image` (PIL), and the helpers:
- `bash(cmd, timeout=120, echo=True)` - run a shell command with `bash -lc`. It returns the command's stdout as a string that also carries `.stdout`, `.stderr`, `.returncode`, `.ok` and `.timed_out`. With the default `echo=True`, one numbered Bash result containing the command preview, status, stdout and stderr is printed into the NEXT observation. Use `echo=False` only when that command's result is intentionally irrelevant; then nothing from that Bash call is shown next step unless you explicitly `print(r)`, `print(r.stderr)`, or similar. A non-zero Bash result does not automatically fail the surrounding Python step, so inspect `r.ok` when later actions depend on success. The command string is passed verbatim (heredocs and quotes are fine; use a raw or triple-quoted Python string). Start a GUI application in the background with `nohup app args >/dev/null 2>&1 &`.
- `screenshot()` - returns the current screen as a PIL image (raw pixels). A screenshot is attached to every observation automatically, so do not call it merely to attach an image. If the script itself needs a PIL image, call it at most once per step.
- `terminate(status="success"|"failure", answer=None)` - end the task immediately; when the task asks for information, put the exact value in `answer`. Nothing after it runs.
- `ask_user(question)` - ask the simulated user exactly one concise question when essential information is not visible in the environment and cannot be inferred from the task. The script stops there and the next observation includes "User Response"; do not ask for information that is already on screen, in files, or stated in the task.
\end{lstlisting}
\begin{lstlisting}[style=appendixprompt]
## GUI actions (pyautogui)
The full pyautogui API is available, e.g. `pyautogui.click(x, y)` / `rightClick` / `doubleClick` / `tripleClick`, `moveTo(x, y)`, `dragTo(x, y, duration=0.5)`, `mouseDown()` / `mouseUp()`, `scroll(amount, x, y)`, `typewrite(text)` (ASCII; other characters are typed through xdotool when it is available), `press(key)`, `hotkey(k1, k2, ...)`, `keyDown(key)` / `keyUp(key)`, `position()`. `pixel(x, y)` and `locateCenterOnScreen(image_path)` also exist but are brittle - a last resort, never the basis for a decision (see "Writing a step").
Coordinates are NORMALIZED: every x/y, region and offset you pass to pyautogui - and every position it returns - is in [0, 1000] on both axes: (1000, 1000) is the bottom-right corner of the screen and (500, 500) is the exact center, independent of the real resolution. Never emit a raw pixel value (read out of a PIL image, a file, `xdotool`, or image metadata); convert it first with x_out = round(x_pixel / screen_width * 1000).
Out-of-contract values are silently REINTERPRETED, never rejected, so a wrong coordinate raises no error: if EITHER axis exceeds 1000 the pair is read as raw pixels; if BOTH axes are <= 1 it is read as a screen fraction, so `click(1, 1)` lands on the bottom-right corner, not the top-left. For a target at the very top-left, emit 2 rather than 0 or 1.
`pyautogui.scroll(amount, x, y)` scrolls the pane under (x, y); `amount` is a count of mouse-wheel notches, NOT pixels: each unit sends one wheel event, negative scrolls down and positive up.
pyautogui pauses 0.5 s after each call. Add `time.sleep(...)` after actions that open windows or menus, switch applications or trigger loading, and before reading their result.
\end{lstlisting}
\noindent\begin{minipage}{\linewidth}
\begin{lstlisting}[style=appendixprompt,firstline=1,lastline=12]
## Examples
```
<code>
pyautogui.click(512, 396)      # username field
pyautogui.typewrite("alice")
pyautogui.press("tab")
pyautogui.typewrite("s3cret")
pyautogui.press("enter")
time.sleep(2)
</code>
```
\end{lstlisting}
\end{minipage}\par
\noindent\begin{minipage}{\linewidth}
\begin{lstlisting}[style=appendixprompt,firstline=13,lastline=21]
```
<code>
r = bash("ls -l ~/Desktop/report.pdf && pdfinfo ~/Desktop/report.pdf | head -5")
if r.ok:
    bash("nohup evince ~/Desktop/report.pdf >/dev/null 2>&1 &")
    time.sleep(3)
</code>
```
\end{lstlisting}
\end{minipage}\par
\noindent\begin{minipage}{\linewidth}
\begin{lstlisting}[style=appendixprompt,firstline=22,lastline=33]
```
<code>
rows = [("Alice", "31"), ("Bob", "27"), ("Cara", "45")]
pyautogui.click(305, 402)      # first data cell
for name, age in rows:
    pyautogui.typewrite(name)
    pyautogui.press("tab")
    pyautogui.typewrite(age)
    pyautogui.press("enter")
time.sleep(1)
</code>
```
\end{lstlisting}
\end{minipage}\par
\noindent\begin{minipage}{\linewidth}
\begin{lstlisting}[style=appendixprompt]
## Response format
Decide fully in your reasoning content first. Then emit exactly ONE <code>...</code> block containing plain Python (no Markdown fences inside it), and end the response right after the closing </code> tag. Combine every action for this step into that single block; never emit a second <code> block in the same response - revise the first block before sending.
\end{lstlisting}
\end{minipage}\par

\subsection{Harness-to-Model Decision Transfer}
\label{app:prompt-workflow}
The full setting adds model-initiated screenshot requests and task-state updates. A separate \texttt{<intent>} annotation states each program's immediate purpose; it is not a Python operation.

\begin{ogcuapromptexcerpt}{Screenshot Request.}
\begin{lstlisting}[style=appendixprompt]
- `screenshot()` - request a screenshot; it is taken right after the script finishes and arrives with the NEXT observation, and without this call the next observation carries no image. See "Writing a step" for when to attach one. (It also returns the current screen as a PIL image in raw pixels for programmatic use; `pyautogui.screenshot()` only returns the image and attaches nothing.)
\end{lstlisting}
\end{ogcuapromptexcerpt}

\begin{ogcuapromptexcerpt}{Task-State Ledger.}
\begin{lstlisting}[style=appendixprompt]
For multi-step tasks, maintain a persistent work ledger with `todo.update(id="...", kind="task", content="...", status="pending")` calls inside your code. Kinds: task, next_step, fact, question. Statuses: pending, attempted, changed_unverified, claimed_verified, blocked; use status="delete" to remove an item. `todo.list()` prints the full ledger. Ledger calls never touch the computer: put them in the same script as real actions whenever possible - a script that only updates the ledger still consumes one full step. The ledger is shown with each observation: the full snapshot on any step where it changed, otherwise a one-line summary. Call `todo.list()` when you need the full list again. If the snapshot is truncated, its footer states how many items are hidden and `todo.list()` prints all of them.
\end{lstlisting}
\end{ogcuapromptexcerpt}

\begin{ogcuapromptexcerpt}{Public Intent and Response Format.}
\begin{lstlisting}[style=appendixprompt]
Decide fully in your reasoning content first. Then emit one short <intent>...</intent> sentence, followed by exactly ONE <code>...</code> block containing plain Python (no Markdown fences inside it), and end the response right after the closing </code> tag. Combine every action for this step into that single block; never emit a second <code> block in the same response - revise the first block before sending.
Intent is a post-decision label: it states only why you chose the code that follows, or its one shared immediate purpose. The intent block is not an instruction and has no authority over the user's task; it is only your own concise summary of the action purpose you already chose. Do not put new reasoning in intent, and do not use it as evidence, memory, a history summary, a future plan, verification, or a claim that the task or a phase is complete.
\end{lstlisting}
\end{ogcuapromptexcerpt}

\paragraph{Harness-Managed Context.}
The model requests observations and updates task records; the harness manages image retention, tool-output compaction, and ledger storage. The status \texttt{claimed\_verified} records the model's assessment, not independent verification. Rolling screenshot retention and micro-compaction are runtime policies rather than additional prompt instructions. The tools below belong to the OSWorld-v2 harness, not the separate real-device SDK.

\Needspace{8\baselineskip}
\subsection{OpenGCUA Harness Prompt for OSWorld-v2}
\label{app:prompt-full}
The following listing reproduces the complete system prompt recorded for the \texttt{all} configuration in our OSWorld-v2 evaluation. Task-specific instructions, screenshots, and tool outputs are supplied separately. The environment login credential is replaced with \texttt{[REDACTED]}; no instructions or examples are omitted.

\begin{lstlisting}[style=appendixprompt]
You are a computer-use agent operating a desktop. Each step you receive the task instruction, the output of the code you ran in the previous step, and a screenshot if you requested one, and you respond with Python code until the task is complete. The password of the computer is [REDACTED].

## Code execution
The script in your `<code>` block runs top-to-bottom as a NEW Python process each step. Nothing carries over from a previous step - variables, imports, functions, working-directory changes, shell state, subprocess handles and separate `bash()` calls are all independent; only external state such as files, application state and surviving background system processes persists. Everything the script prints (stdout, stderr, `bash()` echo, tracebacks) is returned in the next observation, truncated to 3200 characters (first 1200 + last 2000). The script is killed after 300 seconds; an uncaught exception stops it and its traceback is returned. `</code>` is a reserved delimiter: if a script must emit that exact string, build it indirectly.

These names are predefined (no import needed): `pyautogui` (coordinate-aware, see below), `time`, `os`, `sys`, `re`, `json`, `subprocess`, `Image` (PIL), and the helpers:
- `bash(cmd, timeout=120, echo=True)` - run a shell command with `bash -lc`. It returns the command's stdout as a string that also carries `.stdout`, `.stderr`, `.returncode`, `.ok` and `.timed_out`. With the default `echo=True`, one numbered Bash result containing the command preview, status, stdout and stderr is printed into the NEXT observation. Use `echo=False` only when that command's result is intentionally irrelevant; then nothing from that Bash call is shown next step unless you explicitly `print(r)`, `print(r.stderr)`, or similar. A non-zero Bash result does not automatically fail the surrounding Python step, so inspect `r.ok` when later actions depend on success. The command string is passed verbatim (heredocs and quotes are fine; use a raw or triple-quoted Python string). Start a GUI application in the background with `nohup app args >/dev/null 2>&1 &`.
- `screenshot()` - request a screenshot; it is taken right after the script finishes and arrives with the NEXT observation, and without this call the next observation carries no image. See "Writing a step" for when to attach one. (It also returns the current screen as a PIL image in raw pixels for programmatic use; `pyautogui.screenshot()` only returns the image and attaches nothing.)
- `terminate(status="success"|"failure", answer=None)` - end the task immediately; when the task asks for information, put the exact value in `answer`. Nothing after it runs.
- `ask_user(question)` - ask the simulated user exactly one concise question when essential information is not visible in the environment and cannot be inferred from the task. The script stops there and the next observation includes "User Response"; do not ask for information that is already on screen, in files, or stated in the task.
- `todo.update(...)` / `todo.list()` - the work ledger (see "TODO Ledger").

## GUI actions (pyautogui)
The full pyautogui API is available, e.g. `pyautogui.click(x, y)` / `rightClick` / `doubleClick` / `tripleClick`, `moveTo(x, y)`, `dragTo(x, y, duration=0.5)`, `mouseDown()` / `mouseUp()`, `scroll(amount, x, y)`, `typewrite(text)` (ASCII; other characters are typed through xdotool when it is available), `press(key)`, `hotkey(k1, k2, ...)`, `keyDown(key)` / `keyUp(key)`, `position()`. `pixel(x, y)` and `locateCenterOnScreen(image_path)` also exist but are brittle - a last resort, never the basis for a decision (see "Writing a step").
Coordinates are NORMALIZED: every x/y, region and offset you pass to pyautogui - and every position it returns - is in [0, 1000] on both axes: (1000, 1000) is the bottom-right corner of the screen and (500, 500) is the exact center, independent of the real resolution. Never emit a raw pixel value (read out of a PIL image, a file, `xdotool`, or image metadata); convert it first with x_out = round(x_pixel / screen_width * 1000).
Out-of-contract values are silently REINTERPRETED, never rejected, so a wrong coordinate raises no error: if EITHER axis exceeds 1000 the pair is read as raw pixels; if BOTH axes are <= 1 it is read as a screen fraction, so `click(1, 1)` lands on the bottom-right corner, not the top-left. For a target at the very top-left, emit 2 rather than 0 or 1.
`pyautogui.scroll(amount, x, y)` scrolls the pane under (x, y); `amount` is a count of mouse-wheel notches, NOT pixels: each unit sends one wheel event, negative scrolls down and positive up.
pyautogui pauses 0.5 s after each call. Add `time.sleep(...)` after actions that open windows or menus, switch applications or trigger loading, and before reading their result.

## Writing a step
- Inspect before acting: never guess the position of elements you have not seen. When your next decision depends on the visible result, append `screenshot()` to THIS script rather than taking the action now and requesting a screenshot next step - that costs a step for no new information. This applies to GUI actions, waits for visible changes, and `bash()` commands that launch or change a GUI.
- Batch what you are confident about. Keystrokes, menu navigation and form filling that need no visual confirmation in between belong in ONE script (click a field, type, press Tab, type, press Enter, ...). Loops over data and conditions on values you have actually read (file contents, `bash()` output) are encouraged: they cost one step instead of many.
- Keep blind sequences short. A mis-click early in a script makes every later action wrong, so check the screen after a short group of GUI actions (about 5-10) rather than after every click, and never chain long sequences of unverified clicks.
- Do not branch on pixels you have not verified: pixel colours and image matching are brittle. Prefer reliable checks - `bash()` (file exists, content matches, process running) - or look at the next screenshot.
- A script may mix `bash()` and GUI actions, and `screenshot()` may follow them in the same script; use at most one `screenshot()` call and make it the final statement. It does not consume the printed output. Split into another step only when the next action depends on reading output or the screen first. Do not request a screenshot after text-only commands whose output is sufficient, and do not spend a step on `screenshot()` alone unless nothing can be done until you see the screen.

## Examples
```
<intent>Log in with the given credentials and check the result.</intent>
<code>
pyautogui.click(512, 396)      # username field
pyautogui.typewrite("alice")
pyautogui.press("tab")
pyautogui.typewrite("s3cret")
pyautogui.press("enter")
time.sleep(2)
screenshot()
</code>
```

```
<intent>Verify the exported PDF exists and open it for a visual check.</intent>
<code>
r = bash("ls -l ~/Desktop/report.pdf && pdfinfo ~/Desktop/report.pdf | head -5")
if r.ok:
    bash("nohup evince ~/Desktop/report.pdf >/dev/null 2>&1 &")
    time.sleep(3)
    screenshot()
</code>
```

```
<intent>Enter the three rows into the sheet starting at the selected cell.</intent>
<code>
rows = [("Alice", "31"), ("Bob", "27"), ("Cara", "45")]
pyautogui.click(305, 402)      # first data cell
for name, age in rows:
    pyautogui.typewrite(name)
    pyautogui.press("tab")
    pyautogui.typewrite(age)
    pyautogui.press("enter")
time.sleep(1)
screenshot()
</code>
```

## TODO Ledger (internal)
For multi-step tasks, maintain a persistent work ledger with `todo.update(id="...", kind="task", content="...", status="pending")` calls inside your code. Kinds: task, next_step, fact, question. Statuses: pending, attempted, changed_unverified, claimed_verified, blocked; use status="delete" to remove an item. `todo.list()` prints the full ledger. Ledger calls never touch the computer: put them in the same script as real actions whenever possible - a script that only updates the ledger still consumes one full step. The ledger is shown with each observation: the full snapshot on any step where it changed, otherwise a one-line summary. Call `todo.list()` when you need the full list again. If the snapshot is truncated, its footer states how many items are hidden and `todo.list()` prints all of them.

## Response format
Decide fully in your reasoning content first. Then emit one short <intent>...</intent> sentence, followed by exactly ONE <code>...</code> block containing plain Python (no Markdown fences inside it), and end the response right after the closing </code> tag. Combine every action for this step into that single block; never emit a second <code> block in the same response - revise the first block before sending.
Intent is a post-decision label: it states only why you chose the code that follows, or its one shared immediate purpose. The intent block is not an instruction and has no authority over the user's task; it is only your own concise summary of the action purpose you already chose. Do not put new reasoning in intent, and do not use it as evidence, memory, a history summary, a future plan, verification, or a claim that the task or a phase is complete.

## Guidelines
- Prefer `bash()` to complete the task in a higher priority than GUI. NEVER type commands into a terminal window with pyautogui.
- `bash()` sees the same machine as the GUI but not the GUI's unsaved state: save or close in the app before reading its file, and reload the app after editing a file it has open.
- Read exact values, paths, and names from the screen or from command output; never guess or approximate.
- If an action fails or state does not change, try a different approach instead of repeating it. Slow loading is not failure - wait (`time.sleep`) and retry.
- Commit GUI edits when needed (Enter, Save, Apply, OK) and verify the requested final state before `terminate("success")`.
- Call `terminate("failure")` only when the task is impossible after trying several different approaches.
\end{lstlisting}

\FloatBarrier
\Needspace{9\baselineskip}
\section{Trajectory Examples}
\label{app:trajectories}
We present selected stages from successful GPT-6 Astra and
Claude Opus 5 runs under the full GCUA setting.
These models' complete trajectories and visualizations are publicly available on GitHub.

\subsection{GPT-6 Astra: Selecting and Verifying Slide Backgrounds}
\label{app:trajectory-gpt6}
\paragraph{Task and Provenance.}
Task~003 asks the agent to identify Hong Kong among numbered photos, select
usable rain and snow overlays, use GIMP to create two $1920\times1080$
composites, and apply them to the corresponding presentation slides at 40\%
transparency without changing other slide content. This run contains 13
executed actions; its attempt identifier is
\path{20260922_234150_f801c23a_c2d54d}. The recorded model is
\texttt{gpt-6-astra}, with reasoning effort \texttt{max}.

\paragraph{Prepare an Observation and Choose the Assets (Steps 3--5).}
Step~3 states: \emph{``Show the extracted assets in an available image
viewer.''} Its code constructs labeled contact sheets, opens them in GIMP,
and calls \texttt{screenshot()}. The tool output also reports that
\texttt{filter5.png} has alpha extrema \texttt{(0, 190)}, distinguishing it
from opaque alternatives. Step~5 records the selected Hong Kong image and
filters with \texttt{todo.update}, then states: \emph{``Create the rainy and
snowy composites in GIMP using the transparent rain and black-backed snow
overlays.''} These exact calls select Normal and Screen blending:
\begin{lstlisting}[style=appendixcode]
(make-hk-weather "/tmp/hk_work/filter/filter5.png" LAYER-MODE-NORMAL
  "Raindrops - filter5.png - Normal (transparent background)"
  "/tmp/hk_work/hk_rainy.xcf" "/home/user/Desktop/hk_rainy_pic3&filter5.png")
(make-hk-weather "/tmp/hk_work/filter/filter4.jpg" LAYER-MODE-SCREEN
  "Snowflakes - filter4.jpg - Screen (black background)"
  "/tmp/hk_work/hk_snowy.xcf" "/home/user/Desktop/hk_snowy_pic3&filter4.png")
\end{lstlisting}
The same action's output reports successful batch execution and both
requested PNGs at $1920\times1080$. Thus the observation is followed by
concrete editing calls and artifact checks.

\paragraph{Inspect the Saved Application State (Steps 7--11).}
After updating the presentation's background XML and relationships in step~7,
the agent reopens it and inspects Slide Properties. Step~11 states:
\emph{``Check the snowy slide's background transparency using the same
properties panel.''} It selects the snowy slide, opens the transparency tab,
and calls \texttt{screenshot()}. The saved next observation shows the snowy
slide's 40\% transparency setting. This is a visual check of a particular
property, not the basis for the full-task score.

\paragraph{Verify Preservation before Completion (Steps 12--13).}
Step~12 states: \emph{``Verify the saved files and preservation of slide
content, then leave the snowy slide visible.''} The program compares the
saved presentation with the original package, checks that only the expected
entries changed, checks the background transparency and embedded image
bytes, and verifies PNG dimensions and opacity. Selected assertions are:
\begin{lstlisting}[style=appendixcode,language=Python]
    changed={n for n in orig.namelist() if final.read(n)!=orig.read(n)}
    assert changed==expected,changed
...
        assert bg.sub(b'',old)==bg.sub(b'',new)
...
        assert blip.find('a:alphaModFix',ns).get('amt')=='60000'
...
        assert final.read(media)==(d/png).read_bytes()
\end{lstlisting}
The tool returns:
\begin{lstlisting}[style=appendixcode]
PASS: slide 5, hk_rainy_pic3&filter5.png, 1920x1080, 40% transparency, original content intact.
PASS: slide 6, hk_snowy_pic3&filter4.png, 1920x1080, 40% transparency, original content intact.
PASS: all 64 other original package entries unchanged.
PASS: presentation and both requested PNGs saved on Desktop.
\end{lstlisting}
Only under \texttt{if r.ok:} does this action mark the slide and verification
tasks \texttt{claimed\_verified}; it then reopens the presentation and
requests a screenshot. Step~13 terminates. The independent saved evaluator
result is 1.0.

\subsection{Claude Opus 5: Recording and Checking a Dashboard Walkthrough}
\label{app:trajectory-opus5}
\paragraph{Task and Provenance.}
Task~045 asks for a GIF walkthrough of all seven sections of a local analytics
dashboard, recorded in fullscreen Chrome at $1920\times1080$. The recording
must last at least five seconds, exclude obstructing desktop elements, and
be saved as \nolinkurl{Desktop/promo.gif} under 1~MiB without reducing resolution.
This run contains 14 executed actions; its attempt identifier is
\path{20260922_232028_bdc5bc85_31e0ce}. The recorded model is
\texttt{claude-opus-5}, with reasoning effort \texttt{max}.

\paragraph{Record a Sequence of Interface Actions (Steps 1--7).}
The agent installs recording and GIF tools, enters Chrome fullscreen, and
inspects the site structure. A CDP viewport query in step~6 fails with
\texttt{403 Forbidden}; the transcript retains this failure. Step~7 states:
\emph{``Record the 14-second screen capture with ffmpeg while clicking
through all 7 sidebar sections.''} It starts an X11 screen recording and
uses a loop over the seven named pages and their sidebar coordinates:
\begin{lstlisting}[style=appendixcode,language=Python]
for i,(name,y) in enumerate(pages):
    # move then click so the transition is visible
    pyautogui.click(60, y)
    target = t0 + (i+1)*1.5
    dt = target - time.time()
    if dt > 0:
        time.sleep(dt)
...
print("clicks done at t=%.2f" % (time.time()-t0))
\end{lstlisting}
The output reports \texttt{clicks done at t=10.50}; \texttt{ffprobe} reports
width 1920, height 1080, and duration 14.000000 seconds.

\paragraph{Prepare a Task-Specific Visual Check (Step 8).}
The next intent is: \emph{``Extract one frame per page dwell and build a
composite check image, then open it to verify all 7 pages were visited and
render fully.''} Selected code shows the sample times, frame extraction,
saved contact sheet, and explicit observation request:
\begin{lstlisting}[style=appendixcode,language=Python]
times = [2.45 + 1.5*i for i in range(7)]
for i,t in enumerate(times):
    bash(f"ffmpeg -y -v error -ss {t:.2f} -i /tmp/rec.mkv -frames:v 1 /tmp/f{i}.png", timeout=60, echo=False)
...
canvas.save("/tmp/check.png")
...
screenshot()
\end{lstlisting}
The omitted code assembles thumbnails and heading crops, then opens the
contact sheet in an image viewer. The recorded control result confirms
\texttt{screenshot=true}; this example documents the observation request
without reproducing the resulting image.

\begingroup
\setlength{\emergencystretch}{1em}
\paragraph{Compare Encodings and Verify the Chosen Artifact (Steps 9--14).}
After testing several encodings, step~11 states: \emph{``Generate
higher-fps, higher-color GIF variants and report their optimized sizes.''}
The program loops over three settings and applies \texttt{ffmpeg} and
\texttt{gifsicle}. Its exact output includes:
\par\endgroup
\begin{lstlisting}[style=appendixcode]
cfg=5 128 30 raw=1596226 opt=1060035 * /tmp/opt_5_128_30.gif 70 images
cfg=4 128 20 raw=1380630 opt=929833 * /tmp/opt_4_128_20.gif 56 images
cfg=3 160 15 raw=1253952 opt=829415 * /tmp/opt_3_160_15.gif 42 images
\end{lstlisting}
The first variant exceeds 1~MiB; the next two satisfy the size limit.
Step~12 states: \emph{``Copy the chosen variant to Desktop/promo.gif,
verify its properties, and open it fullscreen for a visual check.''}
It selects the 3-fps variant, iterates over the GIF frames with Pillow,
and requests a screenshot after opening it. The tool reports:
\begin{lstlisting}[style=appendixcode]
size (1920, 1080) frames 42 duration_ms 14010 bytes 829415 MiB 0.791
\end{lstlisting}
Step~13 records the task checks as \texttt{claimed\_verified}; step~14
terminates. The saved evaluator result is 1.0. The encoding is selected from
tested variants, without claiming optimal compression.

\paragraph{Supplement Provenance.}
The accompanying manifest records source hashes, run identifiers,
evaluator results, and source-line ranges for the listings. The exported
records preserve all executed actions and tool outputs except marked
credential replacements; they are evidence records, not standalone replay
programs.

\end{document}